\documentclass[10pt,journal,compsoc]{IEEEtran}

\usepackage{ifpdf}
\usepackage{amsmath}

\ifCLASSOPTIONcompsoc
  \usepackage[nocompress]{cite}
\else
  \usepackage{cite}
\fi
\ifCLASSINFOpdf
\else
\fi
\usepackage{amsmath}
\usepackage{algorithmic}

\usepackage{mdwmath}
\usepackage{mdwtab}

\usepackage{stfloats}
\usepackage{url}

\usepackage{amsthm}
\usepackage{amssymb}
\usepackage{multirow}
\usepackage{algorithm}
\usepackage{bm}
\usepackage{subcaption}
\usepackage[colorlinks,linkcolor=red,anchorcolor=magenta,citecolor=red]{hyperref}
\usepackage{xcolor}
\usepackage{csquotes}
\usepackage{adjustbox}
\usepackage{booktabs}
\usepackage{mathtools}
\usepackage{siunitx}

\makeatletter

\newcommand{\Rmnum}[1]{\expandafter\@slowromancap\romannumeral #1@}
\makeatother
\newtheorem{definition}{\textbf{Definition}}
\newtheorem{theorem}{\textbf{Theorem}}
\newtheorem{lemma}{\textbf{Lemma}}
\newtheorem{remark}{\textbf{Remark}}
\newtheorem{assumption}{\textbf{Assumption}}

\begin{document}
%
\title{Learning Non-Vacuous Generalization Bounds from Optimization}
%
%
%
%

\author{Chengli~Tan, Jiangshe~Zhang, Junmin~Liu, 
and~Yihong~Gong, ~\IEEEmembership{Fellow,~IEEE}
\thanks{Chengli Tan is with Northwestern Polytechnical University, Xi'an, China. (Email: cltan023@outlook.com).}
\thanks{Junmin Liu is with Xi'an Jiaotong University, Xi'an, China and SGIT AI Lab, State Grid Corporation of China, Xi'an, China. (Email: junminliu@mail.xjtu.edu.cn)}
\thanks{Jiangshe Zhang and Yihong Gong are with Xi’an Jiaotong University, Xi’an, China. (Email: jszhang, ygong@mail.xjtu.edu.cn)}
}
\markboth{Journal of \LaTeX\ Class Files,~Vol.~14, No.~8, August~2015}%
{Shell \MakeLowercase{\textit{et al.}}: Bare Advanced Demo of IEEEtran.cls for IEEE Computer Society Journals}
%



\IEEEtitleabstractindextext{%
\begin{abstract}
One of the fundamental challenges in the deep learning community is to theoretically understand how well a deep neural network generalizes to unseen data.
However, current approaches often yield generalization bounds that are either too loose to be informative of the true generalization error or only valid to the compressed nets.
In this study, we present a simple yet non-vacuous generalization bound from the optimization perspective.
We achieve this goal by leveraging that the hypothesis set accessed by stochastic gradient algorithms is essentially fractal-like and thus can derive a tighter bound over the algorithm-dependent Rademacher complexity.
The core argument hinges on modeling the discrete-time recursion process via a continuous-time stochastic differential equation driven by fractional Brownian motion.
Numerical studies demonstrate that our approach is able to yield plausible generalization guarantees for modern neural networks such as ResNet and Vision Transformer, even when they are trained on a large-scale dataset (e.g., ImageNet-1K).
\end{abstract}

}

\maketitle

\IEEEdisplaynontitleabstractindextext

%
\IEEEpeerreviewmaketitle

\ifCLASSOPTIONcompsoc
\IEEEraisesectionheading{\section{Introduction}\label{sec:intro}}
\else
\section{Introduction}
\label{sec:intro}
\fi

%
%
%
%
\IEEEPARstart{D}{eep} neural networks (DNNs) have shown remarkable performance in a wide range of tasks over the past decade \cite{bengio2021deep}. 
A mystery is that they generalize surprisingly well on unseen data, though having far more trainable parameters than the number of training examples \cite{belkin2019reconciling, li2023benign}.
This phenomenon of benign overfitting inevitably casts a shadow on
the classical theory of statistical learning, which posits that models with high complexity tend to overfit the training data, whereas models with low complexity tend to underfit the training data.
To reconcile the conflicts, some researchers argue that this is due to the regularization incurred during training, either implicitly imposed via use of stochastic gradient descent (SGD) \cite{advani2020high,barrett2020implicit,smith2020origin, sclocchi2024different} or explicitly via batch normalization \cite{ioffe2015batch}, weight decay \cite{krogh1992simple}, dropout \cite{srivastava2014dropout}, etc.
However, Zhang et al. \cite{zhang2016understanding} questioned this widely received wisdom because they found that DNNs are still able to achieve
zero training error with randomly labeled examples, which apparently cannot generalize.
\par
Prior to our work, there have been extensive studies trying to explain the generalization behavior of DNNs, and they can be roughly categorized into the following classes.
The first class is the so-called norm-based bounds \cite{neyshabur2015norm,bartlett2017spectrally,neyshabur2017pac,golowich2018size} that are composed of the operator norm of layerwise weight matrices.
However, recent studies suggest that these norm-based bounds might be problematic
as they abnormally increase with the number of training examples \cite{nagarajan2019uniform}.
Moreover, norm-based bounds are numerically vacuous as they are even several orders of magnitude larger than the number of network parameters.
The second class connects the generalization to the flatness of the solution \cite{hochreiter1997flat, keskar2016large, dziugaite2017computing,perez2021tighter, nguyen2024flat, zhou2024towards}, showing that flat minima usually generalize well.
However, the flat minima alone do not suffice in explaining the generalization behavior of DNNs. For example, Dinh et al. \cite{dinh2017sharp} argued that sharp minima can generalize as well by reparametrizing the function space, and Wen et al. \cite{wen2024sharpness} also successfully identified a class of non-generalizing flattest models for two-layer ReLU networks.
Another class involves bounding the generalization error via a compression framework \cite{arora2018stronger}.
Empirical results suggest that we can achieve almost non-vacuous bounds on realistic neural networks \cite{zhou2019non,lotfi2022pac}.
Nevertheless, this framework only proves the generalization of the compressed net, not of the true net found by the learning algorithm.
Lastly, stability-based \cite{hardt2016train} and information-theoretic \cite{xu2017information} bounds have also received a lot of attention, but both of them are limited in terms of practical value.
Therefore, it remains a great challenge to search for generalization bounds that not only qualitatively but also quantitatively predict how well the model performs on the new-coming data.
\begin{figure}[t]
    \centering
    \begin{subfigure}[h]{0.24\textwidth}
    \includegraphics[width=\linewidth, clip, trim= 0 0 0 0]{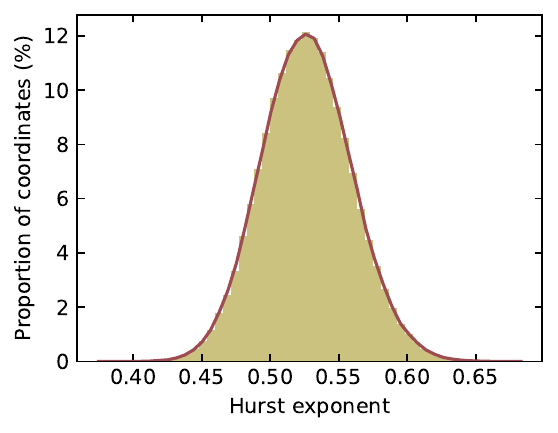}
    \caption{}
    \end{subfigure}
    \begin{subfigure}[h]{0.24\textwidth}
    \includegraphics[width=\linewidth, clip, trim= 0 0 0 0]{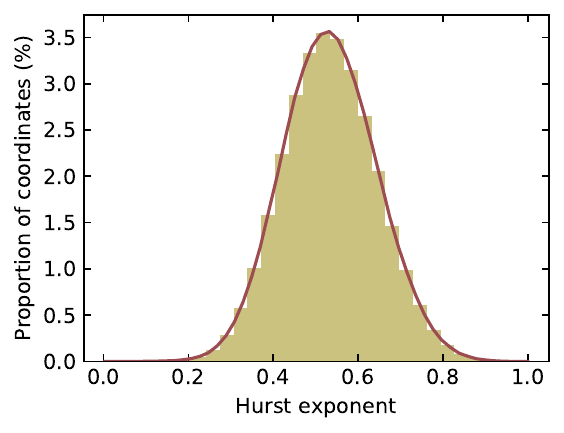}
    \caption{}
    \end{subfigure}
    \caption{Histogram of Hurst exponents for different coordinates of (a) ResNet-56 and (b) NanoGPT. For each coordinate, we first generate a series of stochastic gradient noise (SGN) and then estimate its Hurst exponent.
        If the elements of a time series are mutually independent, for example, in the case of the Brownian motion and the L\'evy flight, the corresponding Hurst exponent would be $1/2$ \cite[Theorem 8.1.3]{embrechts2009selfsimilar}.
        Otherwise, it would suggest that the elements are not independent.
        {Notice that ResNet-56 and NanoGPT are trained by SGD and AdamW, respectively.}
    }
    \label{fig: bulk_of_hurst_exponents}
\end{figure}
\par
Indeed, one critical issue that prevents the generalization bounds from practical usage is that the Rademacher complexity \cite{bartlett2002rademacher} often is evaluated on a pre-specified hypothesis set \cite{neyshabur2015norm,bartlett2017spectrally,arora2019fine}.
But, in practice, we do not want to have a bound that holds uniformly over the pre-specified hypothesis set because we are more interested in a small portion of the hypothesis set that is accessible to the learning algorithm, and our goal is to address this issue. Since most tasks of modern neural networks are attacked by SGD and its variants, we are particularly interested in bounding the Rademacher complexity of the hypothesis set that SGD accesses during training.
\par
To this end, we propose to model the discrete-time SGD recursion through the lens of stochastic differential equations (SDEs), an approach that has been widely used to study the escaping behavior of SGD \cite{jastrzkebski2017three,nguyen2019first,xie2020diffusion}.
An important ingredient to studying SGD from this perspective is stochastic gradient noise (SGN), which is the difference between the stochastic gradient over a mini-batch and the true gradient over the full training set.
In early attempts, by invoking the central limit theorem, SGN is assumed to be either Gaussian \cite{mandt2017stochastic, li2017stochastic, hu2017diffusion, chaudhari2018stochastic,xie2020diffusion, imaizumi2022generalization} or L\'evy stable \cite{simsekli2019tail, zhang2019adam}.
These assumptions are compliant with an implicit constraint that the SGN incurred at different iterations is mutually independent.
However, as shown in Fig. \ref{fig: bulk_of_hurst_exponents}, the temporal correlation of SGN is significant.
Importantly, this correlation is not limited to the classical task of image classification. 
It is also found in modern architectures for natural language modeling.
Therefore, SGN is more reasonable to be fractional Gaussian noise (FGN) rather than Gaussian noise or from the Lévy stable distribution.
Recall that FGNs are the increments of fractional Brownian motion (FBM), a self-similar random process, thus allowing us to quantify the roughness of the optimization trajectory in terms of its Hausdorff dimension.
\par
While the FBM-driven SDE representation of the SGD recursion has previously been investigated \cite{lucchi2022theoretical,tan2021understanding}, they only focused on why SGD favors flat minima, and a rigorous treatment of its relation to generalization is still lacking. 
At the core of our approach lies the fact that the optimization trajectory accessed by SGD during training is restricted to a small subset of the hypothesis space, which is fractal-like due to the incurred FGNs \cite{klingenhofer1999ordinary,lou2016fractal}.
We finally note that there already exist some generalization bounds that take the fractal structure into account, for example, see \cite{simsekli2020hausdorff,camuto2021fractal,dupuis2023generalization,sachs2023generalization,tuci2026generalization}.
However, these approaches only present certain complexity measures, such as the tail index, to compare the generalization performance of one model against that of another.
Neither of them is able to quantitatively give a plausible estimate of the generalization error, and their experimental results are restricted to using a constant learning rate, which is unrealistic for real-world applications.
More seriously, when a classification model is trained with the cross-entropy
loss, Camuto et al. \cite{camuto2021fractal} could not even observe a clear negative or positive correlation between the complexity measure and the generalization error.
By contrast, our approach can yield non-vacuous generalization bounds that predict the test loss well.
Meanwhile, our bound is also model-agnostic, namely, we can efficiently estimate it for any DNNs with complex architectures such as ResNet \cite{he2016deep}, and Vision Transformer \cite{dosovitskiy2020image}. 
\par
The remainder of the paper is organized as follows.
We first review some mathematical notions in Section \ref{sec:preliminaries} and then elaborate on the novel generalization bound for SGD in Section \ref{sec:methodology}. Before concluding, we finally present the experimental results in Section \ref{sec: numberical studies}.

\section{Preliminaries}
\label{sec:preliminaries}
In this section, we briefly recap several concepts that we will use throughout this paper.
\subsection{Fractional Brownian Motion}
\begin{figure*}[t]
    \centering
    \includegraphics[width=0.9\textwidth, clip, trim= 0 0 0 0]{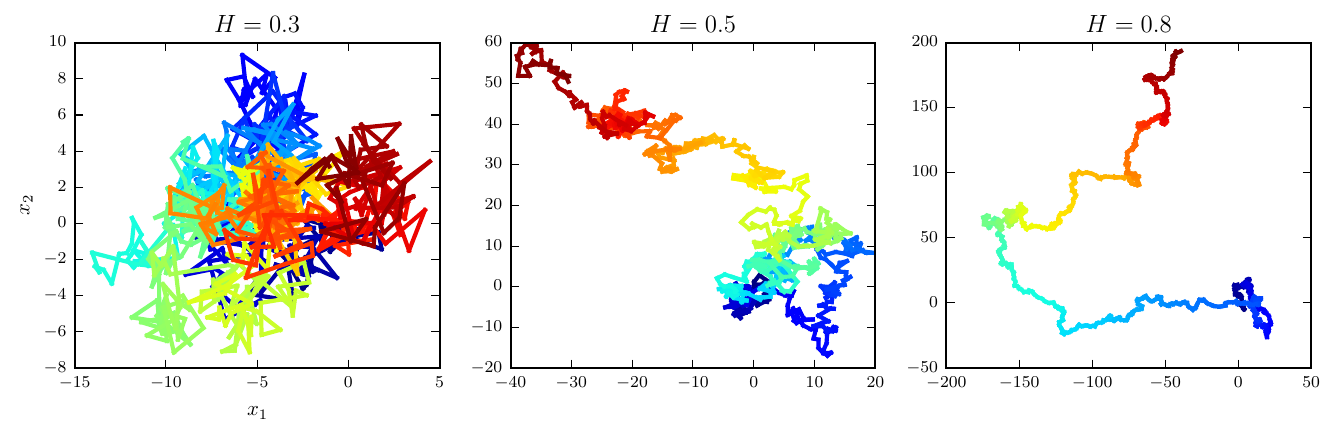}
    \caption{Sample paths of FBM in two-dimensional space. The colors indicate the evolution over time. The Hurst exponent $H$ corresponds to the raggedness of the sample path, with a higher value leading to a smoother motion.}
    \label{fig:sample-path-of-FBM}
\end{figure*}

In probability theory, fractional Brownian motion (FBM), introduced by Mandelbrot and Ness \cite{mandelbrot1968fractional}, is an extension of Brownian motion and is defined as follows.
\begin{definition}
    Given a complete probability space $(\Lambda, \mathcal{B}, \mathbb{P})$,
    FBM is an almost surely continuous centered Gaussian process $\{\Gamma_{H}(t), t\geq 0\}$
    with covariance function
    \begin{equation*}
        \mathbb{E}[\Gamma_{H}(t) \Gamma_{H}(s)] = \frac{1}{2} \left(t^{2H} + s^{2H} - (t-s)^{2H}\right),
    \end{equation*}
    where $H$ is a real value in $(0,1)$ and is often referred to as the Hurst exponent.
\end{definition}
Unlike Brownian motion and other stochastic processes, the increments of FBM need not be independent. In particular, when $H \in (0, 1/2)$, the increments of FBM are negatively correlated and exhibit short-range dependence, implying that it is more likely to overturn past changes. By contrast, FBM shows long-range dependence when $H \in (1/2, 1)$. That is, if it was increasing in the past, it is persistent to keep the trend, and vice versa.
In particular, when $H=1/2$, FBM reduces to the standard Brownian motion.
To gain some intuition,  we plot several sample paths of FBM in Fig. \ref{fig:sample-path-of-FBM} with different Hurst exponents.
One can observe that, when the Hurst exponent $H$ is small,
the sample path is seriously ragged. By contrast, it appears dramatically smoother when the Hurst exponent $H$ becomes relatively larger.
\subsection{Fractal Dimension}
The notion of dimension is central to our analysis.
One that we are most familiar with is the ambient dimension. Roughly speaking, a dimension describes how much space a set occupies near each of its points.
For instance, $\mathbb{R}^d$ as a vector space has an ambient dimension of $d$
since $d$ different coordinates are required to identify a point in this space.
The fractal dimension, however, extends this notion to the fractional case.
While it turns out to be particularly useful in many mathematical fields such as number theory and dynamical systems, there are many different ways to define fractal dimension, and not all the definitions are equivalent to each other.
Of the wide variety of fractal dimensions, we focus on probably the most important box-counting and Hausdorff dimensions.
\par
\textbf{Box-counting dimension}. Suppose $\mathcal{W}$ is a non-empty subset of $\mathbb{R}^d$,
and the diameter of $\mathcal{W}$ is defined as $\mathrm{diam}(\mathcal{W})=\sup\{\|x - y\|: x, y\in \mathcal{W}\}$.
Let $N_\delta(\mathcal{W})$ be the least number of subsets $\{\mathcal{W}_i\}$ of diameter
at most $\delta$ to cover $\mathcal{W}$, that is,
$\mathcal{W}\subseteq \cup_{i=1}^{N_\delta(\mathcal{W})} \mathcal{W}_i$ and $\mathrm{diam}(\mathcal{W}_i)\leq \delta$ for each $i$.
Then, the lower and upper box-counting dimensions of $\mathcal{W}$, respectively,
are defined as
\begin{equation*}
    \underline{\dim}_{\text{box}} \mathcal{W}=\varliminf_{{\delta \rightarrow 0}} \frac{\log N_{\delta}(\mathcal{W})}{\log(1/\delta)},
\end{equation*}
and
\begin{equation*}
    \overline{\dim}_{\text{box}} \mathcal{W}=\varlimsup_{{\delta \rightarrow 0}} \frac{\log N_{\delta}(\mathcal{W})}{\log(1/\delta)}.
\end{equation*}
Note that
$\underline{\dim}_{\text{box}} \mathcal{W} \le \overline{\dim}_{\text{box}} \mathcal{{W}}$
and if the equality holds, the box-counting dimension of $\mathcal{W}$
is then denoted by
\begin{equation*}
    \dim_{\text{box}} \mathcal{W}=\lim_{{\delta \rightarrow 0}} \frac{\log N_{\delta}(\mathcal{W})}{\log(1/\delta)}.
\end{equation*}
The popularity of the box-counting dimension is largely due to its intuitive definition and relative ease of empirical calculation.
By contrast, the Hausdorff dimension, which is described below, is in terms of measure theory and is mathematically convenient to work with.
Consequently, a disadvantage of the Hausdorff dimension is that it is often difficult to estimate by computational methods.
However, for a proper understanding of fractal geometry, familiarity with the Hausdorff dimension is essential.
\par
\textbf{Hausdorff dimension}. Let $\{\mathcal{W}_i\}_{i=1}^{\infty}$ be a $\delta$-cover of a non-empty bounded set $\mathcal{W}$,
and for each ${\alpha}\ge0$, we call
\begin{equation*}
    \resizebox{0.99\hsize}{!}{$\mathfrak{H}^{\alpha}(\mathcal{W})=\lim_{{\delta \rightarrow 0}}\left\{\sum_{i=1}^{\infty}\mathrm{diam}(\mathcal{W}_i)^{\alpha}: \mathcal{W}\subseteq \cup_{i=1}^{\infty}\mathcal{W}_i, \mathrm{diam}(\mathcal{W}_i)<\delta\right\},$}
\end{equation*}
the $\alpha$-dimensional Hausdorff measure of $\mathcal{W}$.
Usually, it equals $0$ or $\infty$.
The critical value of $\alpha$ at which $\mathfrak{H}^\alpha(\mathcal{W})$
jumps from $\infty$ to $0$ is referred to as the Hausdorff dimension.
Rigorously, it is defined as
\begin{align*}
    \dim_{\text{haus}} \mathcal{W}&=\inf \left\{\alpha \geq 0: \mathfrak{H}^{\alpha}(\mathcal{W})=0\right\}\\
    &=\sup \left\{\alpha\geq 0: \mathfrak{H}^{\alpha}(\mathcal{W})=\infty\right\}.
\end{align*}
\noindent
While these two kinds of dimensions are the same under some regularity conditions \cite[Theorem 5.7]{mattila1999geometry}, they are not equivalent to each other.
For example, considering the set of rationals in $[0, 1]$, the Hausdorff dimension is 0, while the box-counting dimension is 1. 
In general, however, it holds that $\dim_{\text{haus}} \mathcal{W}\leq\dim_{\text{box}} \mathcal{W}$.

\section{Methodology}
\label{sec:methodology}
Assume we have access to a training set of independent and identically distributed (\emph{i.i.d.}) data points, 
\begin{equation*}
    S=\{(x_1, y_1), \ldots, (x_m, y_m)\}= \{z_1, \cdots, z_m\},
\end{equation*}
where $x\in\mathcal{X}$ denotes the features, $y\in\mathcal{Y}$ denotes the labels, and $\mathcal{Z}=\mathcal{X}\times \mathcal{Y}$ denotes the data space that follows an unknown data distribution $\mathfrak{D}$.
The goal of supervised learning is to choose a suitable hypothesis $f_w:\mathcal{X}\mapsto\mathcal{Y}$, parameterized by a vector of network parameters $w\in\mathbb{R}^d$, so that the generalization error (\emph{i.e.} the risk on previously unseen data),
\begin{equation*}
    R_{\mathfrak{D}}(w)= \underset{z\sim \mathfrak{D}}{\mathbb{E}}\left[\ell(w, z)\right] = \underset{(x, y)\sim \mathfrak{D}}{\mathbb{E}}\left[\mathcal{L}(f_w(x), y)\right]
\end{equation*}
is small.
Here, $\mathcal{L}: \mathcal{Y}\times\mathcal{Y}\mapsto \mathbb{R}_{+}$ is a non-negative loss function, and $\ell: \mathbb{R}^d\times\mathcal{Z} \to \mathbb{R}_{+}$ is the composition of the loss and the hypothesis, which will also referred to as \enquote{loss}, with a slight abuse of notation.
\par
However, due to the unknown data distribution $\mathfrak{D}$, we are not able to minimize $R_{\mathfrak{D}}(w)$ directly.
Instead, we can only minimize the empirical error over the training set $S$, namely,
\begin{equation*}
    R_S(w) = \frac{1}{m} \sum_{i=1}^{m} \ell(w, z_i).
\end{equation*}
Notice that the difference $R_{\mathfrak{D}}(w)-R_S(w)$ is referred to as the generalization gap. 
Particularly, in the realizable case where the empirical error is zero, the generalization gap is interchangeable with the generalization error.
It should be stressed that a small generalization gap does not necessarily dictate a small generalization error (requiring the training loss to be small as well).
For example, for an untrained neural network, the generalization gap between the training set and the test set is small, whereas the generalization error on the test set could be very large.
\subsection{Problem Setup}
Starting from an initialization point $w_0\in\mathbb{R}^d$, the SGD algorithm recursively updates the weights of the neural network as follows,
\begin{equation}
    w_{k+1} = w_k - \eta \nabla\hat{\ell}_{S_k}(w_k),
    \label{eq:sgd-recursion}
\end{equation}
where $\eta$ is the learning rate and $\nabla\hat{\ell}_{S_k}(w_k)$ is an unbiased estimate of the true gradient, which is computed by
\begin{equation*}
    \nabla\hat{\ell}_{S_k}(w_k)=\frac{1}{|S_k|} \sum_{z \in S_k} \nabla \ell(w_k, z),
\end{equation*}
where $S_k$ is a set of examples (\emph{i.e.} mini-batch) that are \emph{i.i.d.}~drawn from $S$ and $b=|S_k|$ is the mini-batch size.
Particularly, when $S_k = S$, SGD becomes the full-batch gradient descent (GD).
While the SGD algorithm is random, once the training set $S$, the initialization point $w_0$, and the training steps $K$ are fixed, the total number of optimization trajectories (\emph{i.e.} the collection of weights throughout training) $M(m, b, K)$ is indeed finite (though very large).
To see this, notice that there are only finitely many subsets that $S_k$ can take.
For example, in the case of with-replacement sampling, there are in total $m^b$ mini-batches to choose from at every step.
By contrast, in the case of without-replacement sampling, this number can be further reduced to $m \choose b$.
Of course, here we require that there are no other sources of stochasticity during training, such as perturbing the weights with random noise.
\par
Many studies \cite{zhu2018anisotropic,amir2021sgd,wu2023implicit} have shown that training neural networks with the stochastic gradient $\nabla\hat{\ell}_{S_k}(w_k)$ generally outperforms with the true gradient $\nabla\hat{\ell}_{S}(w_k)$ because of the incurred stochastic gradient noise (SGN), which is defined as
\begin{equation*}
    \zeta_k=\nabla\hat{\ell}_{S_k}(w_k)-\nabla\hat{\ell}_{S}(w_k).
\end{equation*}
If one assumes that the learning rate $\eta$ is sufficiently small and $\zeta_k$ follows a zero-mean distribution, the SGD recursion \eqref{eq:sgd-recursion} can be seen as a first-order discretization of a continuous-time SDE \cite{li2017stochastic}.
Perspectives from SDEs have provided many insights on studying the generalization behavior of DNNs through the asymptotic convergence rate and local dynamic behavior of SGD \cite{mandt2017stochastic,simsekli2019tail,xie2020diffusion,tan2021understanding,gess2024stochastic}.
Recall that Fig.\ref{fig: bulk_of_hurst_exponents} shows that it is more reasonable to model the driving noise with FGN.
Therefore, in our analysis, we will consider the case where SGD is viewed as the Euler-Maruyama discretization of the following SDE,
\begin{equation}
    \mathrm{d}w_t = -\mu(w_t, t)\mathrm{d}t + \sigma(w_t, t)\mathrm{d} \Gamma_{H}(t),
    \label{eq:fbm-sde}
\end{equation}
where $\mu(w_t, t)\in\mathbb{R}^d$ is the drift coefficient, $\sigma(w_t, t)\in\mathbb{R}^d$ is the diffusion coefficient, and $\Gamma_H(t)$ represents a $d$-dimensional FBM with Hurst exponents $H=(H_1, \ldots, H_d)$.
Such a class of SDEs admits SGN produced at different iterations to be mutually interdependent, which significantly varies from previous studies where SGN is assumed either to be Gaussian \cite{mandt2017stochastic, li2019stochastic} or follow a L\'evy stable distribution \cite{simsekli2020hausdorff,dupuis2024generalization}.
\par
\begin{figure}[t]
    \centering
    \begin{subfigure}[h]{0.45\linewidth}
    \centering
    \includegraphics[width=\linewidth]{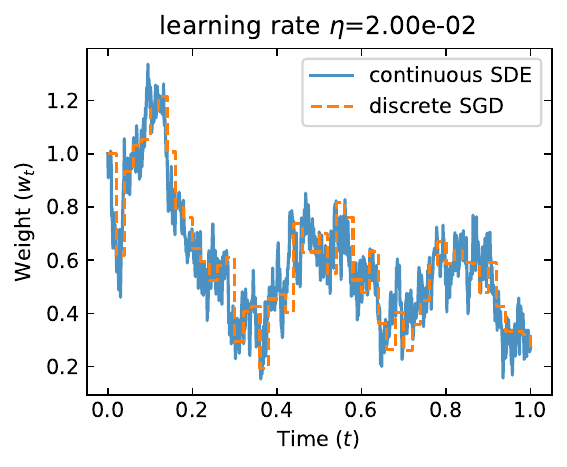}
    \caption{$H=0.3$}
  \end{subfigure}\hfill
  \begin{subfigure}[h]{0.45\linewidth}
    \centering
    \includegraphics[width=\linewidth]{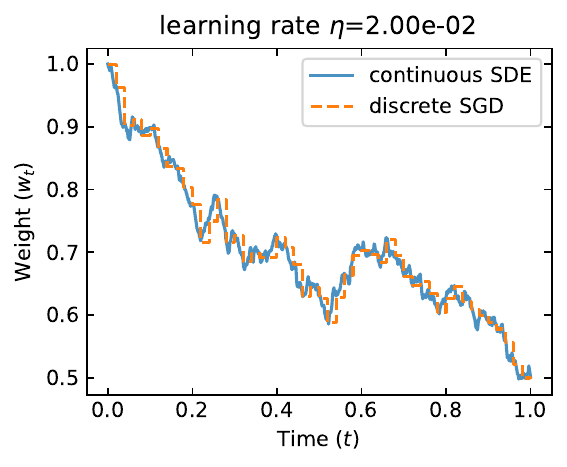}
    \caption{$H=0.7$}
  \end{subfigure}\\
  \begin{subfigure}[h]{0.45\linewidth}
    \centering
    \includegraphics[width=\linewidth]{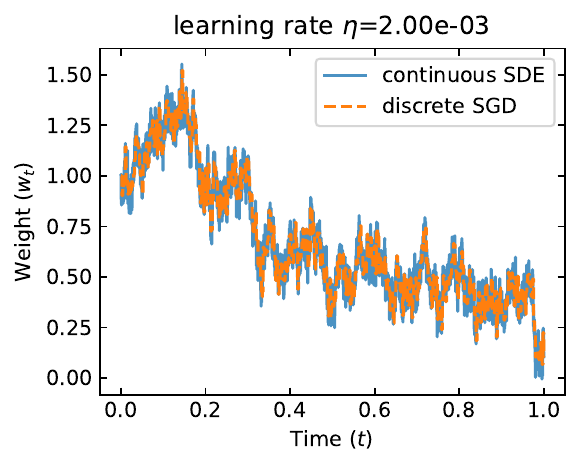}
    \caption{$H=0.3$}
  \end{subfigure}\hfill
  \begin{subfigure}[h]{0.45\linewidth}
    \centering
    \includegraphics[width=\linewidth]{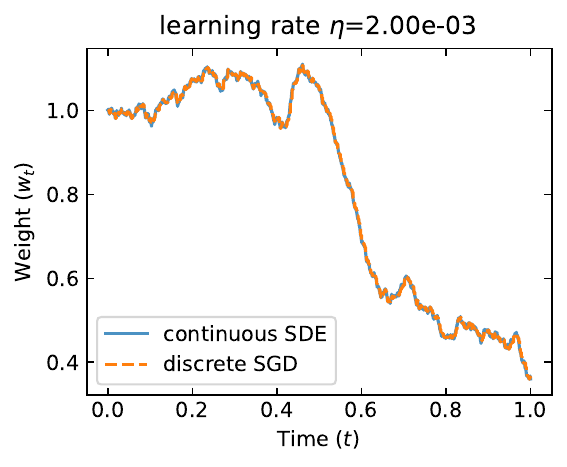}
    \caption{$H=0.7$}
  \end{subfigure}
    \caption{Illustration of the coupling between continuous-SDE and discrete-SGD for the fractional Ornstein-Uhlenbeck process.
    }
    \label{fig: sample path sgd vs sde}
\end{figure}

Indeed, a pairwise correspondence between discrete-time SGD recursion~\eqref{eq:sgd-recursion} and continuous-time SDE driven by FBM~\eqref{eq:fbm-sde} can be easily established.
For a finite number $K$ of training steps, when the learning rate $\eta$ is small enough,
for any $t \in [k\eta, (k+1)\eta)$, we can always define a stochastic process,
\begin{equation*}
    \widehat{w}_t = w_k - (t - k\eta)\mu(w_k, k\eta) + \sigma(w_k, k\eta)(\Gamma_H(t) - \Gamma_H(k\eta)),
\end{equation*}
as the interpolation of two successive iterates $w_k$ and $w_{k+1}$ such that $\widehat{w}_{k\eta}=w_k$ for all $k\in\{0, 1, \cdots, K\}$.
This approach is frequently adopted in SDE literature \cite{mishura2008rate} and allows the trajectory to be continuous to represent the SGD recursion.
For simplicity, we also assume that the random noise of different coordinates is mutually independent.
Moreover, when the diffusion coefficient $\sigma(w_t, t)$ is constant, the following theorem establishes the strong coupling bound between the discrete sequence $w_k$ of Eq.\eqref{eq:sgd-recursion} and the continuous SDE solution $w_t$ of Eq.\eqref{eq:fbm-sde}, confirming that our continuous approximation holds in the small learning rate regime.
\begin{theorem}
\label{theorem: convergence of sgd to sde}
Let $w_t$ be the solution to the SDE,
\begin{equation}
\label{eq: solution to fbm sde}
    w_t = w_0 - \int_0^t \mu(w_s, s) \mathrm{d}s + \sigma \Gamma_H(t),
\end{equation}
and let $w_k$ be the discrete SGD iterates with learning rate $\eta$,
\begin{equation*}
    w_{k+1} = w_k - \eta \mu(w_k, t_k) + \sigma (\Gamma_H(t_{k+1}) - \Gamma_H(t_k)),
\end{equation*}
where $t_k = k\eta, k\in\{0, 1, \cdots, K\}$. Under the assumption that $\mu (w_t, t)$ is Lipschitz continuous, for any $0<\alpha <H$, there always exists some constant $C>0$ such that the  error at time $T=K\eta$ satisfies 
\begin{equation*}
    \mathbb{E}[|w_K - w_T|] \leq C \eta^\alpha.
\end{equation*}
\end{theorem}
\begin{remark}
    Since Theorem \ref{theorem: convergence of sgd to sde} holds for any $\alpha < H$, the convergence order is effectively $H$.
    That is, the discretization error is smaller when the Hurst exponent $H$ is larger. 
\end{remark}
  To illustrate how the SDE couples with the discrete optimization steps, we further provide a simple toy example, namely, the fractional Ornstein-Uhlenbeck process.
  Consider the following SDE,
  \begin{equation}
    \label{eq: fou process}
      \mathrm{d}w_t = -\theta w_t \mathrm{d}t + \sigma\mathrm{d} \Gamma_H(t),
  \end{equation}
  where $-\theta w_t$ is the drift term and $\sigma\mathrm{d} \Gamma_H(t)$ is the fractional noise with Hurst exponent $H$.
  This SDE has an analytical solution and is given by
  \begin{equation*}
      w_T = e^{-\theta T} w_0 + \sigma \int_0^T e^{-\theta (T-s)} \mathrm{d}\Gamma_H(s).
  \end{equation*}
  And the discrete version (Euler-Maruyama) of SDE \eqref{eq: fou process} with learning rate $\eta$ is
  \begin{equation*}
      w_{k+1} = w_k - \theta w_k \eta + \sigma (\Gamma_H(t_{k+1}) - \Gamma_H(t_k)),
  \end{equation*}
which is exactly the update rule for SGD on a quadratic loss $\ell(w) = \theta w^2 / 2$.
Figure \ref{fig: sample path sgd vs sde} depicts how the red \enquote{stairs} (SGD) follow the blue curve (SDE).
As the learning rate $\eta \to 0$, the \enquote{height} of these stairs shrinks and the SDE is the mathematical limit of the SGD updates. 
Moreover, the discretization error is larger when the Hurst exponent $H$ is smaller, which is consistent with the result of Theorem \ref{theorem: convergence of sgd to sde}.
\par
Therefore, the SGD optimization trajectory $W^{ 0:K}_{\xi|S, w_0}=\left\{w_0, w_1, \ldots, w_K\right\}$, indexed by $\xi\in\{1, \ldots, M(m, b, K)\}$, always can be viewed as a sample path of the solution to SDE \eqref{eq:fbm-sde} in a time frame, say,  without loss of generality,  $w([0, T]) = \{w_t, t\in[0, T]\}$.
Consequently, for a training set $S$ and an initialization point $w_0$, the hypothesis set $\mathcal{W}_{S, w_0}$ that SGD can access is the collection of $M(m, b, K)$ such kind of sample paths and is essentially a tiny space \cite{gur2018gradient,li2022low}.
\par
While $w_0$ is randomly drawn from a probability distribution, unless otherwise specified, our discussion below always assumes that $w_0$ is fixed so that our analysis can be greatly simplified.
This is because most SGD solutions trained from different initialization points belong to the same basin in the loss landscape after proper permutation \cite{entezari2022role,ainsworth2023git}.
As a result, any generalization bounds conditioned on $w_0$ can also be applied to predict the generalization performance of  SGD solutions that are trained from another initialization point.
For simplicity of notation, we will omit the dependence on $w_0$ and simply write $\mathcal{W}_S$ instead.
Further, we write $\mathcal{G}_S$ to denote the loss functions associated with $\mathcal{W}_S$ mapping from $\mathcal{Z}=\mathcal{X}\times\mathcal{Y}$ to $\mathbb{R}_{+}$,
\begin{equation*}
    \mathcal{G}_S = \{g_w = \ell(w, z) | w\in \mathcal{W}_S \}.
\end{equation*}
To remove the dependence on $S$, we can take a union over $S\in\mathcal{Z}^m$, yielding $\mathcal{W} = \cup_{S\in \mathcal{Z}^m} \mathcal{W}_S $ and $\mathcal{G} = \cup_{S\in \mathcal{Z}^m} \mathcal{G}_S$ to represent the set of all possible parameters and loss functions.
For any $\varepsilon>0$, our goal is to bound the following term
\begin{equation*}
    \mathbb{P}\left[\sup_{w\in \mathcal{W}}|\widehat{R}_S(w) - R(w)|\geq\varepsilon\right],
\end{equation*}
which is algorithm-dependent and differs from what is usually studied, where $\mathcal{W}$ is replaced by a pre-specified hypothesis set.
In the sequel, we will present the main result in terms of the empirical Rademacher complexity $\mathfrak{R}_S(\mathcal{G})$ \cite{bartlett2002rademacher}, which is defined as
\begin{align*}
    \mathfrak{R}_S(\mathcal{G}) &= \frac{1}{m}~ \underset{\boldsymbol{\sigma} \sim\{\pm 1\}^{m}}{\mathbb{E}}\left[\sup _{g_w \in \mathcal{G}} \sum_{i=1}^{m} \sigma_{i}g_w(z_i)\right] \\
    &= \frac{1}{m}~ \underset{\boldsymbol{\sigma} \sim\{\pm 1\}^{m}}{\mathbb{E}}\left[\sup _{w \in \mathcal{W}} \sum_{i=1}^{m} \sigma_{i}~ \ell(w, z_i)\right],
\end{align*}
where the Rademacher variables $\sigma_i$ are \emph{i.i.d.}~with $\mathbb{P}(\sigma_i=\pm 1)=1/2$.
Let $\mathcal{H}=\mathcal{G} \circ S$ be the set of all possible loss evaluations that a loss function $g_w\in \mathcal{G}$ can achieve over the training set $S$, namely,
\begin{align*}
    \mathcal{H}= \mathcal{G} \circ S = \left\{h_w = \left(g_w(z_1), \ldots, g_w(z_m)\right) | g_w\in \mathcal{G}\right\}.
\end{align*}
We can further observe that the value of $\mathfrak{R}_S(\mathcal{G}) $ is the same as the Rademacher complexity $\mathfrak{R}(\mathcal{H})$ of the set $\mathcal{H} \subset \mathbb{R}_{+}^m$.
\par
In the following section, we aim to control $\mathfrak{R}(\mathcal{H})$ by taking into account the Hausdorff dimension of the sample paths of the solution to SDE \eqref{eq:fbm-sde}.
The Hausdorff dimension determines the raggedness of the sample path and characterizes the dynamic behavior of SGD around the local minimum.
%
%
\subsection{Upper Bound}
We will first present several assumptions used in our theoretical analysis.
\begin{assumption}
    \label{assumption: lipschitz continuous}
    The loss function $\ell: \mathbb{R}^d\times\mathcal{Z}\mapsto\mathbb{R}_+$ is bounded in $[0, 1]$ and there exists a constant $L \geq 1$ such that the loss function is Lipschitz continuous with respect to its first argument.
\end{assumption}
The boundedness assumption is standard in the literature, for example, see \cite{shalev2014understanding,mohri2018foundations}.
Furthermore, if a mapping satisfies the Lipschitz continuity,
then the Hausdorff dimension of the image is no greater than the Hausdorff
dimension of the preimage \cite[Proposition 3.3]{falconer2004fractal}.
This Lipschitz assumption can be easily satisfied if the gradient of the loss function is uniformly bounded for any $w\in\mathbb{R}^d$, for example, by gradient clipping.
\begin{assumption}
    \label{assumption: exsitence}
    The drift coefficient $\mu(w_t, t)$ and diffusion coefficient $\sigma(w_t, t)$ in SDE \eqref{eq:fbm-sde} are both bounded vector fields on $\mathbb{R}^d$.
\end{assumption}
This assumption is reasonable due to the
existence of batch normalization \cite{ioffe2015batch}, weight decay \cite{krogh1992simple}, and other popular tricks. Under this assumption, the existence and uniqueness of solutions to SDE \eqref{eq:fbm-sde} are guaranteed if the Hurst exponent $H$ is larger than $1/4$ \cite{lyons2002system}.
For the case of $H<1/4$, however, there is no similar theoretical result so far.
Since most of the Hurst exponents are larger than $1/4$, as shown in Fig.\ref{fig: bulk_of_hurst_exponents}, this assumption suffices to ensure the existence and uniqueness of the solutions in our analysis.
\begin{assumption}
    \label{assumption: regularity}
    Let $\mathcal{W}$ be a non-empty bounded subset of $\mathbb{R}^d$ and there exists a Borel measure $\nu$ on $\mathbb{R}^d$ and positive numbers $a$, $b$, $r_0$ and $\kappa$ such that $0 < \nu(\mathcal{W}) \le \nu(\mathbb{R}^d) < \infty$ and for $w\in \mathcal{W}$
    \begin{equation*}
        0<a r^{\kappa} \leq \nu(B(w, r)) \leq b r^{\kappa}<\infty, \quad 0<r<r_0,
    \end{equation*}
    where
    \begin{equation*}
        B(w, r) = \{w^\prime\in\mathcal{W}|\|w - w^\prime\|< r\}.
    \end{equation*}
\end{assumption}
This so-called Ahlfors regularity is often used in fractal geometry to ensure the set is regular enough so that the Hausdorff dimension is equivalent to the box-counting dimension \cite[Theorem 5.7]{mattila1999geometry}.
That is, under this assumption, we have $\dim_{\text{box}}\mathcal{W}=\dim_{\text{haus}}\mathcal{W}=\kappa$.
As a result, we can use the covering number
techniques.
Recall that $\mathcal{W}$ is a collection of sample paths of the solution to SDE \eqref{eq:fbm-sde}, and thus we have $\kappa\geq 1$ as well.
\par
Based on these assumptions, we are ready to present an upper bound over $\mathfrak{R}(\mathcal{H})$. 
\begin{lemma}
    \label{theorem:single-trajectory}
    Let Assumptions \ref{assumption: lipschitz continuous}-\ref{assumption: regularity} hold. For any \emph{i.i.d.}~sample $S\in\mathcal{Z}^m$, there always exist a constant $c\geq 1$ such that the following inequality holds:
    \begin{equation*}
        			 \mathfrak{R}(\mathcal{H})\leq \frac{6\mathrm{diam}(\mathcal{H})}{m}\sqrt{2\dim_{\text{haus}}\mathcal{W}}\left(\hat{\beta}^{-1/2} + \hat{\beta}^{1/2}\right),
        \label{eq:bound-single}
    \end{equation*}
    		where $\hat{\beta} = \log {c\sqrt{m}L}/{\mathrm{diam}(\mathcal{H})}$.
\end{lemma}
Based on the Rademacher complexity $\mathfrak{R}(\mathcal{H})$, we are now ready to present the bound over the maximal generalization gap.
\begin{theorem}
    \label{theorem:main}
    Let Assumptions \ref{assumption: lipschitz continuous}-\ref{assumption: regularity} hold.
    Then, for any $\tau>0$, with probability at least $1-\tau$ over the draw of an \emph{i.i.d.}~sample $S\in\mathcal{Z}^m$, there always exists a constant $c\geq 1$ such that the following inequality holds for all $w \in \mathcal{W}$,
    		\begin{align*}
        			&R_{\mathfrak{D}}(w) - R_S(w) \\
           &\leq  \frac{12\mathrm{diam}(\mathcal{H})}{m} \sqrt{2\dim_{\text{haus}}\mathcal{W}} \left(\hat{\beta}^{-1/2} + \hat{\beta}^{1/2}\right) + 3\sqrt{\frac{1}{2m}\log{\frac{2}{\tau}}},
        		\end{align*}
\end{theorem}
\begin{proof}
    This is a direct consequence of \cite[Theorem 3.3]{mohri2018foundations}.
\end{proof}
\begin{remark}
    For the simplest case where $\hat{\beta} = 1$, the above bound reduces to $\mathcal{O}({\mathrm{diam}(\mathcal{H})\sqrt{\dim_{\text{haus}}\mathcal{W}}}/{m})$,
    indicating that the generalization gap continues to grow until the training process saturates because $\mathrm{diam}(\mathcal{H})$ is always increasing.
    This observation is consistent with the results of algorithmic stability \cite{hardt2016train}.
    Moreover, notice that our bound does not explicitly depend on the number of trainable parameters $d$.
    Instead, the Hausdorff dimension $\dim_{\text{haus}} \mathcal{W}$ plays a similar role and quantifies the \enquote{effective} complexity of the hypothesis set because $\dim_{\text{haus}} \mathcal{W}$ in general is much smaller than $d$.
\end{remark}
\begin{remark}
    In the classical literature where the fractal structure of the learned hypothesis set is not taken into consideration, the Rademacher complexity $\mathfrak{R}(\mathcal{H})$ scales as $\mathcal{O}(\sqrt{\log m})$ if we assume  $\operatorname{diam}(\mathcal{H}) \propto \sqrt{m}$, see \cite[Example 27.2]{shalev2014understanding}.
    As a result, this suggests that the generalization bound would increase with the number of training examples, which is obviously contradictory to the empirical results.
    By contrast, our result suggests that the above bound can decrease with the number of training examples at a sublinear rate, namely, $\mathcal{O}(1/\sqrt{m})$.
\end{remark}
\subsection{Estimation}
While Theorem \ref{theorem:main} provides the structural components of the generalization bound, it remains challenging to compute in practice.
Particularly, we do not know how to derive $\dim_{\text{haus}} \mathcal{W}$.
In the following, we will first present a relaxed version of Theorem \ref{theorem:main} so that $\dim_{\text{haus}} \mathcal{W}$ can be efficiently estimated.
\begin{assumption}
    \label{assumption:finite countability}
    The data distribution $\mathfrak{D}$ is supported on a countable set.
\end{assumption}
This assumption generally holds for image- and text-based datasets since each image pixel is an integer from $0$ to $255$ and the sentences are segmented into a vocabulary of tokens.
We note that the countability assumption is crucial to our results.
Thanks to this condition, we are able to invoke the countable stability \cite[Section 3.2]{falconer2004fractal} of the Hausdorff dimension to control the upper bound of $\dim_\text{haus} \mathcal{W}$,
\begin{align*}
     \dim_{\text{haus}} \mathcal{W} &= \dim_{\text{haus}} \cup_{S\in \mathcal{Z}^m} \mathcal{W}_S = \sup_{S\in\mathcal{Z}^m} \dim_{\text{haus}} \mathcal{W}_S.
\end{align*}
Recall that $\mathcal{W}_S$ can be viewed as a collection of $M(m, b, K)$ sample paths of the solution to SDE \eqref{eq:fbm-sde}. Therefore, its Hausdorff dimension is the Hausdorff dimension of the sample paths of the solution to SDE \eqref{eq:fbm-sde}.
\par
\begin{figure}[t]
    \centering
    \begin{subfigure}[h]{0.15\textwidth}
        \includegraphics[width=\linewidth, clip, trim= 0 0 0 0]{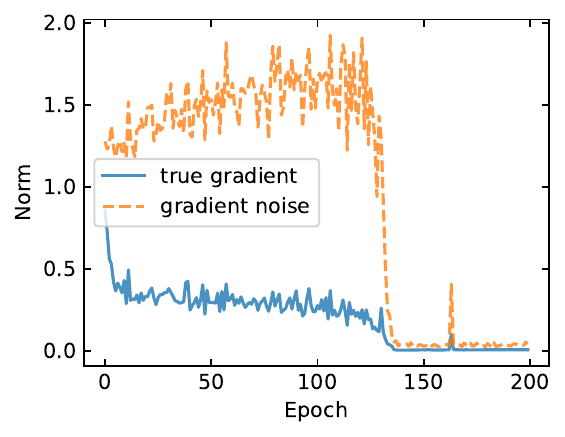}
        \caption{}
    \end{subfigure}
    \begin{subfigure}[h]{0.15\textwidth}
        \includegraphics[width=\linewidth, clip, trim= 0 0 0 0]{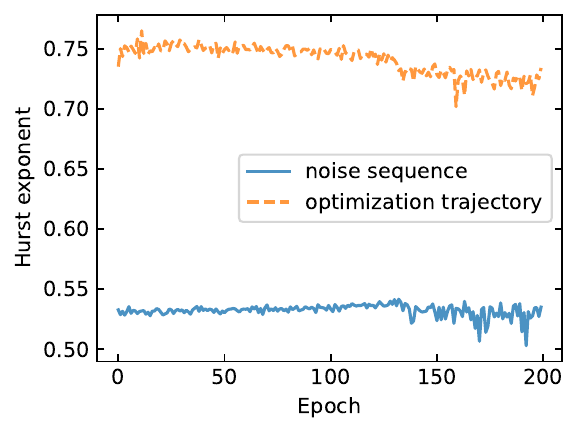}
        \caption{}
    \end{subfigure}
    \begin{subfigure}[h]{0.15\textwidth}
        \includegraphics[width=\linewidth, clip, trim= 0 0 0 0]{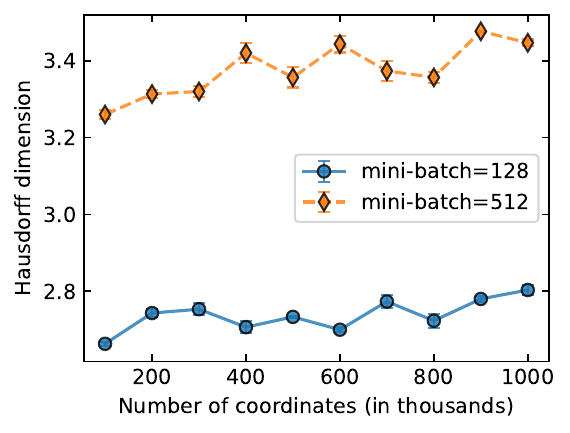}
        \caption{}
    \end{subfigure}
    \caption{(a) Norm of the true gradient and the stochastic gradient as a function of training epoch. 
    (b) Evolution of the average Hurst exponent over the course of training.
    Notice that for each coordinate, the Hurst exponent of SGN is evaluated on a sequence of length 2000 by backpropagating the gradients with different mini-batches.
    In contrast, the Hurst exponent of the optimization trajectory is evaluated on the collected parameters by consecutively updating the model 2000 steps with a small learning rate.
    (c) Hausdorff dimension $\dim_{\text{haus}} \Gamma_{H}([0, T])$ estimated using different numbers of coordinates of the neural network.}
    \label{fig:number coordinates}
\end{figure}

However, the current study on the Hausdorff dimension of the sample paths of the solution to SDE \eqref{eq:fbm-sde} is only limited to the case where the Hurst exponent $H$ is the same for all coordinates \cite{lou2016fractal}.
This obviously is not true for real-world neural networks that have millions (even billions) of parameters (see Fig. \ref{fig: bulk_of_hurst_exponents}).
Therefore, it remains unknown how to derive $\dim_{\text{haus}} \mathcal{W}_S$.
Luckily, under certain conditions, the Hausdorff dimension of the sample path to the solution of SDE \eqref{eq:fbm-sde} $\dim_{\text{haus}}(w([0,T]))$ can be controlled by the Hausdorff dimension of the driving noise $\dim_{\text{haus}}(\Gamma_H([0,T]))$.
\begin{assumption}
\label{lemma: bound over traj dimension}
        Let $w_t$ be the solution to the SDE \eqref{eq:fbm-sde} driven by a mutually independent multi-Hurst FBM $\Gamma_H(t)$. Assume that the drift and the diffusion coefficients are sufficiently smooth so that the Hausdorff dimension of the sample path almost surely satisfies 
        \begin{equation*}
            \dim_{\mathrm{haus}}(w([0,T])) \leq \dim_{\mathrm{haus}}(\Gamma_H([0,T])).
        \end{equation*}
\end{assumption}
 Empirically, as shown in Fig. \ref{fig:number coordinates}(a), the norm of SGN is always much larger than the norm of the true gradient, suggesting that the training process is primarily dominated by the diffusion term.
 On the other hand, as shown in Fig. \ref{fig:number coordinates}(b), we can observe that the Hurst exponent estimated from SGN is consistently larger than that estimated from the optimization trajectory.
 This implies that the optimization trajectory is smoother than the sample path of the driving FBM, and thus, this assumption is reasonable.
 Moreover, the Hurst exponent remains almost constant over the course of training, suggesting that the driving noise is stationary and can be safely used to estimate $\dim_{\text{haus}} \Gamma_{H}([0, T])$.
\par
By imposing this assumption, we can use the known results of multi-dimensional FBM.
According to Remark 2 of \cite{yimin1994dimension} and Theorem 2.1 of \cite{xiao1995dimension}, the Hausdorff dimension of the sample path of the multi-Hurst FBM, with probability 1, is
\begin{equation}
    \label{eq: estimate hausdorff dimension}
    \dim_{\text{haus}} \Gamma_{H}([0, T]) = \frac{1 + \sum_{i=1}^{k}(H_k - H_i)}{H_k}, 
\end{equation}
where the Hurst exponents are sorted such that $0<H_1\leq H_2\leq \cdots \leq H_d <1$ and $k$ is determined by the inequality $\sum_{i=1}^{k-1}H_i \leq 1 \leq \sum_{i=1}^{k}H_i$.
\par
To remove the supremum over the random sampling of $S\in\mathcal{Z}^m$, we need to introduce another assumption on the driving noise.
\begin{assumption}
    \label{assumption: hurst independence}
    The Hurst exponent $H$ does not depend on the \emph{i.i.d} sampling process of $S$ from the data distribution $\mathfrak{D}$, and the order of the elements in the training set $S$.
\end{assumption}
By imposing this condition, we mean that the Hurst parameter $H$ is determined solely by the network architecture and the optimizer hyperparameters such as the learning rate and mini-batch size.
It remains unchanged when the neural network receives different training examples from the same data distribution.
Consequently, the Hausdorff dimension $\dim_{\text{haus}} \Gamma_{H}([0, T])$ corresponding to the driven FBM does not depend on the order of the mini-batches. Namely, for any specific run of SGD, it remains the same.
This can be easily checked by shuffling the order of mini-batches (see Table \ref{tab:stochasticity on hausdorff dimension}).
Furthermore,  we can also observe that $\dim_{\text{haus}} \Gamma_{H}([0, T])$ remains approximately the same even when the model is trained with different training sets and initialization points.
Therefore, the Hausdorff dimension $\dim_{\text{haus}} \Gamma_{H}([0, T])$ estimated under any specific run of SGD essentially provides a plausible upper bound over  $\dim_{\text{haus}} \mathcal{W}$, which is particularly useful in practice.
\begin{theorem}
    \label{theorem:estimation}
    Let Assumptions \ref{assumption: lipschitz continuous}-\ref{assumption: hurst independence} hold.
    Then, for any $\tau>0$, with probability at least $1-\tau$ over the draw of an \emph{i.i.d.}~sample $S\in\mathcal{Z}^m$, there always exists a constant $c\geq 1$ such that the following inequality holds for all $w \in \mathcal{W}$,
    		\begin{align*}
        			&R_{\mathfrak{D}}(w) - R_S(w) - 3\sqrt{\frac{1}{2m}\log{\frac{2}{\tau}}} \\
           &\leq  \frac{12\mathrm{diam}(\mathcal{H})}{m} \sqrt{2\dim_{\mathrm{haus}} \Gamma_{H}([0, T])} \left(\hat{\beta}^{-1/2} + \hat{\beta}^{1/2}\right),
        		\end{align*}
    where $\hat{\beta} = \log {c\sqrt{m}L}/{\mathrm{diam}(\mathcal{H})}$.
\end{theorem}
\begin{proof}
    This is a direct consequence of Theorem \ref{theorem:main}.
\end{proof}
While the generalization bound in Theorem \ref{theorem:estimation} is not as tight as the one in Theorem \ref{theorem:main}, it is numerically computable even for large neural networks.
In the sequel, we will use $\varrho_{\mathrm{bound}}$ as an estimator of the theoretical generalization bound, which is defined as
\begin{table}[t]
    \centering
    \caption{Effects of different sources of stochasticity on Hausdorff dimension $\dim_{\text{haus}} \Gamma_{H}([0, T])$.
        The first row quantifies how $\dim_{\text{haus}} \Gamma_{H}([0, T])$ is affected by the different initialization points of the neural network (ResNet-20) under the same training set.
        When the neural network is initialized with the same weights, the second row describes how $\dim_{\text{haus}} \Gamma_{H}([0, T])$ changes with the training set (\emph{i.e.} random subsets of CIFAR-10).
        Finally, when both the initialization point and the training set are the same, the last row further studies the effect of the order of the mini-batches.
    }
    \resizebox{0.48\textwidth}{!}{
    \begin{tabular}{@{}lcccc@{}}
        \toprule
        & \multicolumn{4}{c}{Number of training examples (per class)}                            \\ \cmidrule{2-5}
        & 1000 & 2000  & 3000 & 4000\\ \cmidrule{1-5}
        Initialization point               & 3.12 $\pm$ 0.07        & 2.84 $\pm$ 0.09  & 2.79 $\pm$ 0.07 & 2.70 $\pm$ 0.06     \\
        Training set               & 3.02 $\pm$ 0.09       & 2.84 $\pm$ 0.06 & 2.78 $\pm$ 0.04  & 2.71 $\pm$ 0.04        \\
        Mini-batch order               & 3.03 $\pm$ 0.09         & 2.83 $\pm$ 0.04 & 2.77 $\pm$ 0.02 & 2.71 $\pm$ 0.04        \\\bottomrule
    \end{tabular}
    }
    \label{tab:stochasticity on hausdorff dimension}
\end{table}
\begin{equation*}
    \varrho_{\mathrm{bound}}= \frac{12\mathrm{diam}(\mathcal{H})}{m} \sqrt{2\dim_{\text{haus}} \Gamma_{H}([0, T])} \left(\beta^{-1/2} + \beta^{1/2}\right),
\end{equation*}
where $\beta=\log {\sqrt{m}L}/{\mathrm{diam}(\mathcal{H})}$.
Compared to Theorem \ref{theorem:estimation}, notice that $\beta$ omits the nuisance factor $\log c$ of $\hat{\beta}$ because it is essentially an artifact due to the proof.
Indeed, if $\mathcal{{W}}$ is a self-similar set or generated from an iterated function system \cite{falconer2004fractal,camuto2021fractal}, the value of $c$ approximately equals to $1$.
Apart from the already known number of training examples $m$, there are three remaining terms to be estimated:
\par
(1) Approximation of Lipschitz constant $L$.
    The theoretical $L$ represents the global Lipschitz constant of the loss functions.
Although we have assumed $L$ as a constant that universally holds for any $w, w^\prime \in\mathbb{R}^d$, in practice, it should be restricted to the space of $\mathcal{W}$ and therefore corresponds to a much smaller value.
Recall that the Lipschitz continuity can be guaranteed if the gradient of the loss function is bounded, namely, $\|\nabla \ell_w(w, z)\| \leq L$ for any $w\in\mathcal{W}$ and $z\in \mathcal{Z}$.
Moreover,  we have at each iteration $\operatorname{Var}(\nabla \ell_{w_k}(w_k, z)) =|S_k| \operatorname{Var}(\nabla\hat{\ell}_{S_k}(w_k))$.
Therefore, we can approximate $L$ with the maximum value of  $|S_k|^{1/2}\|\nabla\hat{\ell}_{S_k}(w_k)\|$ throughout training.
\par
(2) Approximation of diameter $\mathrm{diam}(\mathcal{H})$.
The true diameter $\mathrm{diam}(\mathcal{H})$ measures the maximum distance between any two models in the loss vector space.
To this end, we need to calculate the per-example loss on the full training set until the training is finished.
Subsequently, we can estimate the diameter of $\mathcal{H}\subset \mathbb{R}_+^m$ by computing the smallest bounding ball \footnote{Code is available at \url{https://github.com/hirsch-lab/cyminiball}.}.
However, this approach is computationally prohibitive when $m$ is very large.
To circumvent this issue, we can alternatively approximate $\mathrm{diam}(\mathcal{H})$ with 
\begin{equation*}
    \resizebox{0.98\hsize}{!}{$\left[\left(\ell(w_K, z_1) - \ell(w_0, z_1)\right)^2 + \cdots + \left(\ell(w_K, z_m) - \ell(w_0, z_m)\right)^2 \right]^{\frac{1}{2}},$}
\end{equation*}
where $w_0$ and $w_K$ are the vectors of network parameters at initialization and the end of training.
This is because the loss is always non-negative and generally tends to decrease during training.
\par
(3) Approximation of $\dim_{\text{haus}} \Gamma_{H}([0, T])$.
According to Eq. \eqref{eq: estimate hausdorff dimension}, we are able to give an estimation of the Hausdorff dimension, for which we first need to estimate the Hurst exponent \footnote{Code is available at \url{https://github.com/CSchoel/nolds}.} for each coordinate of the neural network. 
To produce a series of SGN for a neural network, we run through the full training set to calculate the full-batch gradient.
Then, we feed a number of mini-batches into the neural network, and as a result, we can obtain a series of SGN by subtracting the full-batch gradient from the mini-batch gradient.
Notice that for very large neural networks that contain millions (even billions) of trainable parameters, due to limited memory, we are not able to generate a series of SGN for each coordinate.
In this case, we can randomly sample a small portion of coordinates, and we find that the estimation is robust to the number of used coordinates (see Fig. \ref{fig:number coordinates}b).
\par
Finally, we want to emphasize that these terms, theoretically, should be better estimated using the union of multiple runs with different seeds.
In practice, however, we find that they often lead to similar results. 
Therefore, we choose to estimate $\varrho_{\mathrm{bound}}$ using a single run, which is particularly useful in scenarios such as neural architecture search where an instant measure is required to compare against different runs. 
\section{Numerical Studies}
\label{sec: numberical studies}
In this section, we present the experimental results to demonstrate the efficacy of the proposed generalization bound.
The code to reproduce the results is available at \url{https://github.com/aiu-group/bound2024}.
\subsection{Implementation Details}
Here we consider three publicly available datasets---CIFAR-10, CIFAR-100 \cite{krizhevsky2009learning}, and ImageNet-1K \cite{deng2009imagenet}.
CIFAR-10 and CIFAR-100 are composed of $50,000$ training examples and $10,000$ test examples that are equally divided into 10 and 100 classes.
By contrast, ImageNet-1K is a large-scale dataset that consists of 1000 classes and contains approximately one million training images and $50,000$ validation images.
We do not use data augmentation in all experiments,
since doing so will prevent the model from consistently reaching low
cross-entropy loss and impose uncontrollable effects on SGN as the training examples are no longer \emph{i.i.d.}~distributed \cite{dziugaite2020search, jiang2019fantastic}.
\par
Unless otherwise specified, optimization uses SGD with momentum of $0.9$ and weight decay of \num{5.0e-4}.
By default, we use a mini-batch size of $128$, a learning rate of $0.05$, and sometimes a cosine learning rate scheduler to ensure that the models can fit the training set completely.
Determining when to stop the training process is important to quantitatively
assess the generalization bounds, especially for those that can only be calculated after the training is finished.
Stopping too early or too late may produce different results.
Slightly different from \cite{jiang2019fantastic, dziugaite2020search}, we terminate the training process when the training accuracy reaches the threshold of $99.5$\%.
This is because decreasing the cross-entropy loss to a very low value will result in severe overfitting.
\subsection{Number of Training Examples}
\label{subsection: number of training examples}
Increasing the number of training examples generally will promote the generalization performance of DNNs \cite{kaplan2020scaling}.
While this observation is obvious, a non-negligible fact is that there are still a large number of generalization bounds that fail to (correctly) reveal this correlation \cite{nagarajan2019uniform}. 
\par
In the following, we aim to investigate how the proposed bound $\varrho_{\mathrm{bound}}$ changes with the number of training examples. 
First, we need to generate a bunch of subsets as follows: for CIFAR-10, we gradually increase the number of training examples (per class) from 500 to 5000 with a step size of 500; and for CIFAR-100, the number is increased from 100 to 500 with a step size of 50.
We then proceed to train two modern neural networks---ResNet-56 \cite{he2016deep}, and WideResNet-28-10 \cite{zagoruyko2016wide}---for 50 and 200 epochs, respectively.
\par
\begin{figure}[t]
    \centering
    \begin{subfigure}[h]{0.24\textwidth}
    \includegraphics[width=\linewidth, clip, trim= 0 0 0 0]{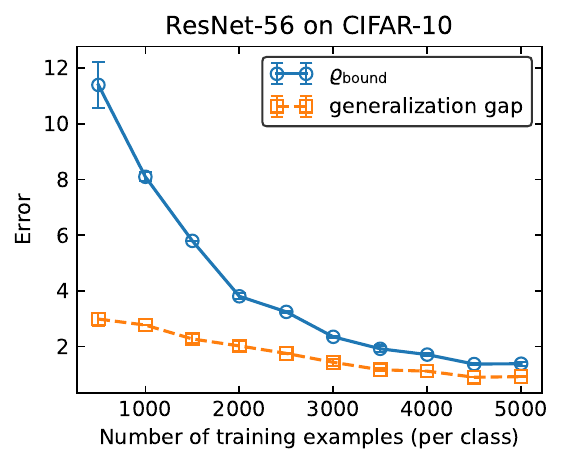}
    \end{subfigure}
    \begin{subfigure}[h]{0.24\textwidth}
    \includegraphics[width=\linewidth, clip, trim= 0 0 0 0]{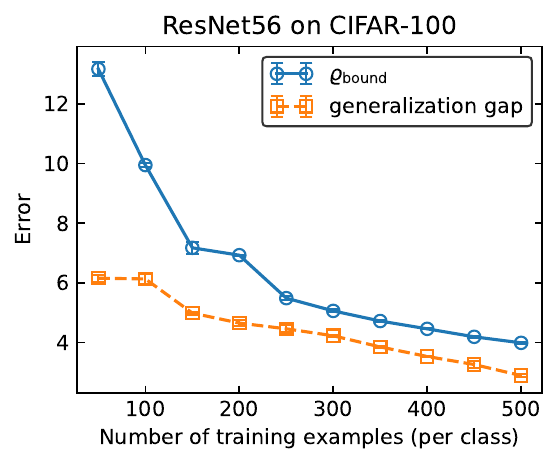}
    \end{subfigure} \\
    \begin{subfigure}[h]{0.24\textwidth}
    \includegraphics[width=\linewidth, clip, trim= 0 0 0 0]{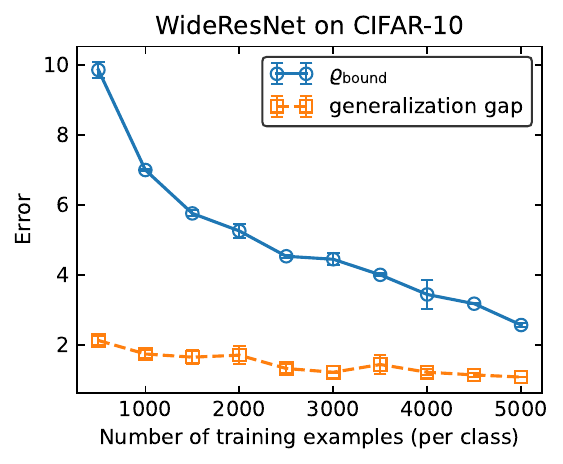}
    \end{subfigure}
    \begin{subfigure}[h]{0.24\textwidth}
    \includegraphics[width=\linewidth, clip, trim= 0 0 0 0]{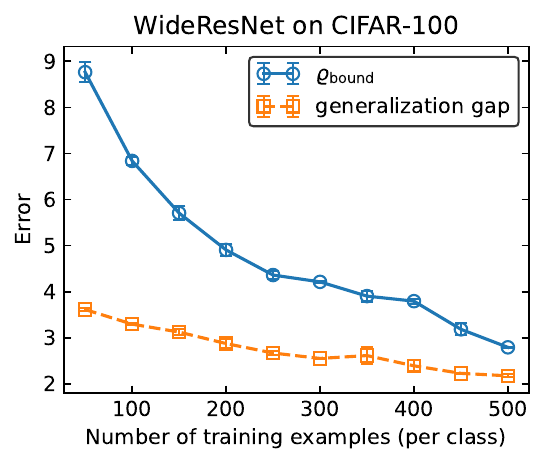}
    \end{subfigure} \\
    \caption{Upper bound $\varrho_{\mathrm{bound}}$ and generalization gap (test loss-training loss) as a function of the number of training examples. Note the learning rate remains constant over the course of training, and the momentum coefficient is zero.
    }
    \label{fig: number_of_training_examples_const}
\end{figure}
\begin{figure}[t]
    \centering
    \begin{subfigure}[h]{0.24\textwidth}
    \includegraphics[width=\linewidth, clip, trim= 0 0 0 0]{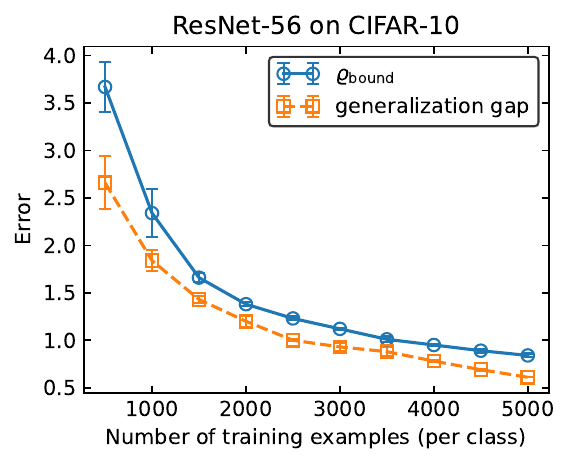}
    \end{subfigure}
    \begin{subfigure}[h]{0.24\textwidth}
    \includegraphics[width=\linewidth, clip, trim= 0 0 0 0]{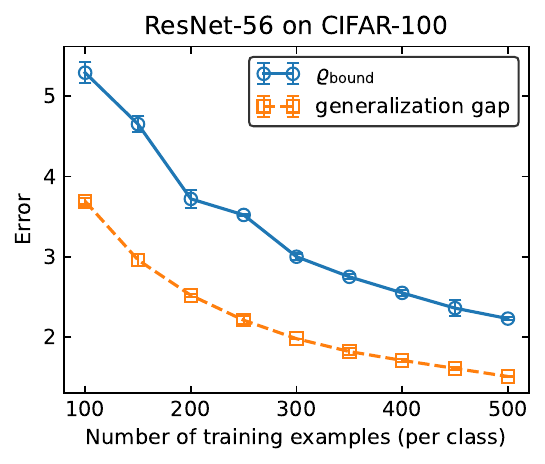}
    \end{subfigure} \\
    \begin{subfigure}[h]{0.24\textwidth}
    \includegraphics[width=\linewidth, clip, trim= 0 0 0 0]{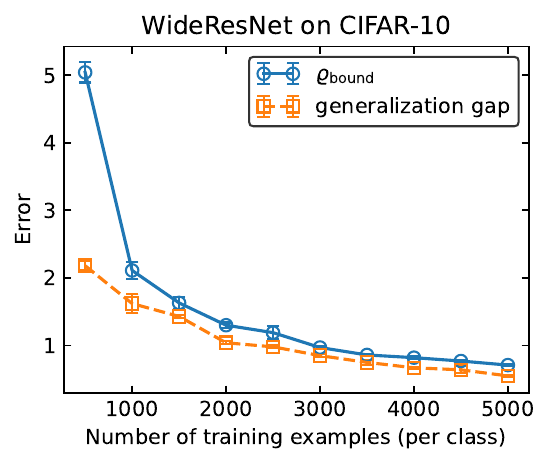}
    \end{subfigure}
    \begin{subfigure}[h]{0.24\textwidth}
    \includegraphics[width=\linewidth, clip, trim= 0 0 0 0]{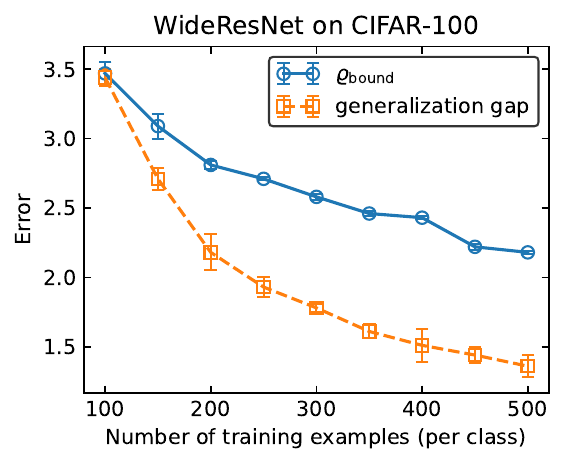}
    \end{subfigure} \\
    \caption{Upper bound $\varrho_{\mathrm{bound}}$ and generalization gap (test loss-training loss) as a function of the number of training examples.
    Note that the learning rate is decayed to zero according to a cosine scheduler over the course of training.
    }
    \label{fig: number_of_training_examples}
\end{figure}

In Figs. \ref{fig: number_of_training_examples_const} and \ref{fig: number_of_training_examples}, we illustrate the results for constant and annealed learning rate, respectively.
The generalization gap (test loss - training loss) indeed decreases as more training examples are used, and our bound $\varrho_{\mathrm{bound}}$ correctly captures this trend.
More importantly, we observe that $\varrho_{\mathrm{bound}}$ is non-vacuous and can almost recover the generalization gap on CIFAR-10 when the full training set is used.
\subsection{Effects of Learning Rate and Mini-batch Size}
Another issue that hinders previous generalization bounds from wide usage is that they often anti-correlate
with the generalization error when changing the commonly used training hyperparameters \cite{jiang2019fantastic}.
In this part, we aim to probe the effects of learning rate and mini-batch size,
which typically dominate the generalization performance of DNNs.
To this end, we varied the learning rate from $0.02$ to $0.1$ with a step size of $0.02$ and simultaneously doubled the mini-batch size from $64$ to $1024$.
\par
\begin{figure}[t]
    \centering
    \begin{subfigure}[h]{0.24\textwidth}
    \includegraphics[width=\linewidth, clip, trim= 0 0 0 0]{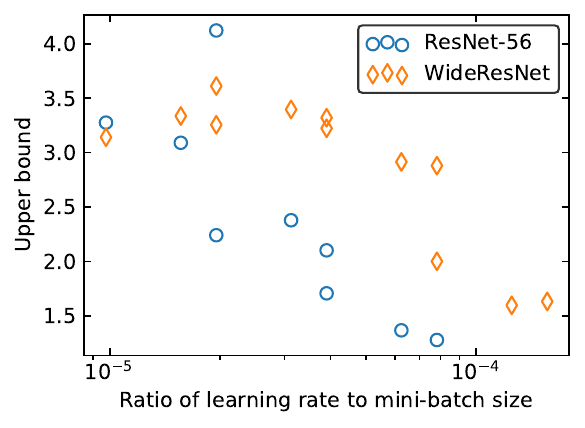}
    \caption{CIFAR-10}
    \end{subfigure}
    \begin{subfigure}[h]{0.24\textwidth}
    \includegraphics[width=\linewidth, clip, trim= 0 0 0 0]{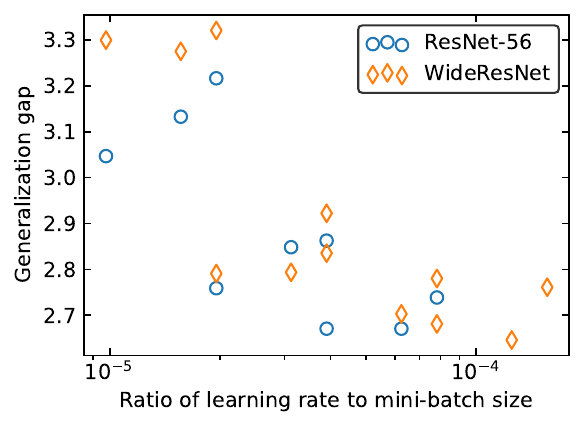}
    \caption{CIFAR-10}
    \end{subfigure} \\
    \begin{subfigure}[h]{0.24\textwidth}
    \includegraphics[width=\linewidth, clip, trim= 0 0 0 0]{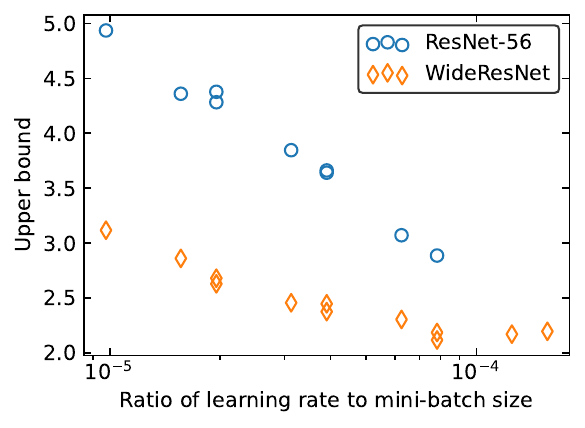}
    \caption{CIFAR-100}
    \end{subfigure}
    \begin{subfigure}[h]{0.24\textwidth}
    \includegraphics[width=\linewidth, clip, trim= 0 0 0 0]{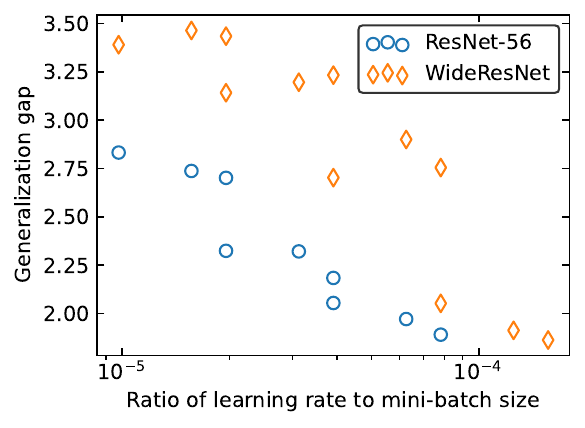}
    \caption{CIFAR-100}
    \end{subfigure} \\
     \caption{Negative correlation between the upper bound $\varrho_{\mathrm{bound}}$ and the ratio of learning rate to mini-batch size. Note the learning rate remains constant over the course of training, and the momentum coefficient is zero.}
    \label{fig:learning rate and batch size const}
\end{figure}
\begin{figure}[t]
    \centering
    \begin{subfigure}[h]{0.24\textwidth}
    \includegraphics[width=\linewidth, clip, trim= 0 0 0 0]{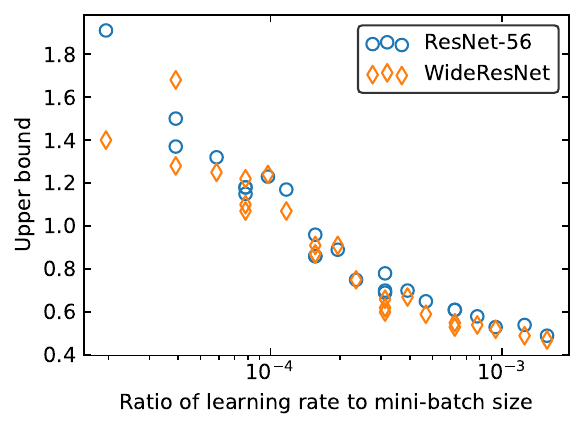}
    \caption{CIFAR-10}
    \end{subfigure}
    \begin{subfigure}[h]{0.24\textwidth}
    \includegraphics[width=\linewidth, clip, trim= 0 0 0 0]{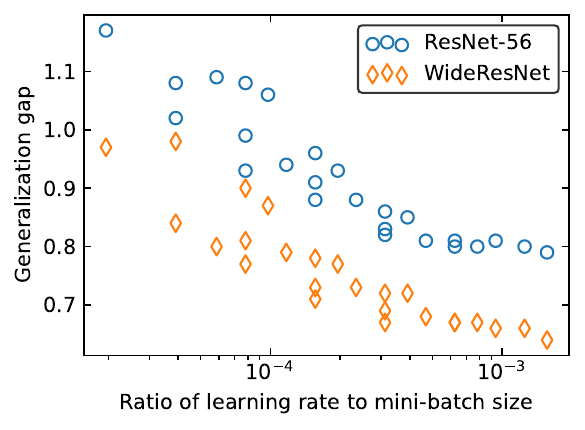}
    \caption{CIFAR-10}
    \end{subfigure} \\
    \begin{subfigure}[h]{0.24\textwidth}
    \includegraphics[width=\linewidth, clip, trim= 0 0 0 0]{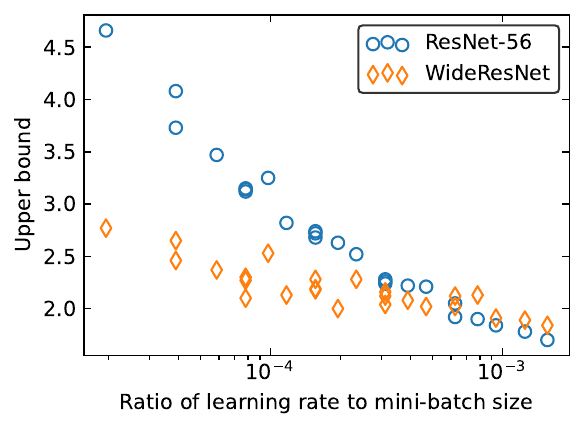}
    \caption{CIFAR-100}
    \end{subfigure}
    \begin{subfigure}[h]{0.24\textwidth}
    \includegraphics[width=\linewidth, clip, trim= 0 0 0 0]{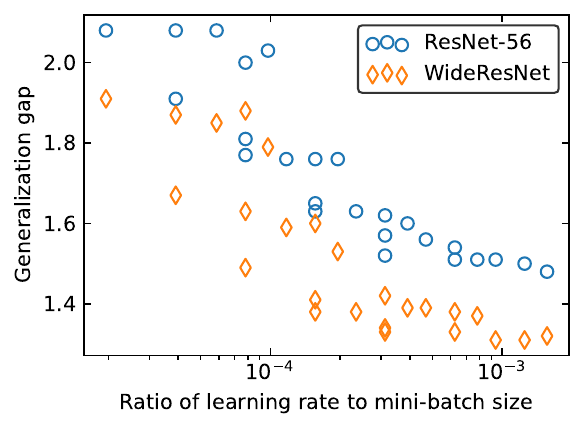}
    \caption{CIFAR-100}
    \end{subfigure} \\
     \caption{Negative correlation between the upper bound $\varrho_{\mathrm{bound}}$ and  the ratio of learning rate to mini-batch size. Note that the learning rate is decayed to zero according to a cosine scheduler over the course of training.}
    \label{fig:learning rate and batch size}
\end{figure}

As shown in Figs. \ref{fig:learning rate and batch size const} and \ref{fig:learning rate and batch size}, we can observe that the upper bound $\varrho_{\mathrm{bound}}$ indeed decreases with the ratio of the learning rate to the mini-batch size.
These results align with the observation that a larger ratio of learning rate to mini-batch size usually leads to a better generalization \cite{jastrzkebski2017three,he2019control}.


\subsection{Results on ImageNet-1K}
In this section, we continue to investigate how the proposed bound $\varrho_{\mathrm{bound}}$ evolves with the training epoch.
Particularly, we evaluate it on the large-scale ImageNet-1K dataset.
For this purpose, we trained on two popular neural networks---ResNet-18 and ViT-S-32 \cite{dosovitskiy2020image}---with basic data augmentation, namely, resizing and cropping images to 224-pixel resolution and then normalizing them.
Both models are trained for 300 epochs to ensure that the neural networks are fully trained.
The optimizer is AdamW with an initial learning rate of 3.0e-3 and a weight decay of 0.1.
For both models, a cosine schedule is used to adjust the learning rate.
\par
\begin{figure}[t]
    \centering
    \begin{subfigure}[h]{0.15\textwidth}
        \includegraphics[width=\linewidth, clip, trim= 0 0 0 0]{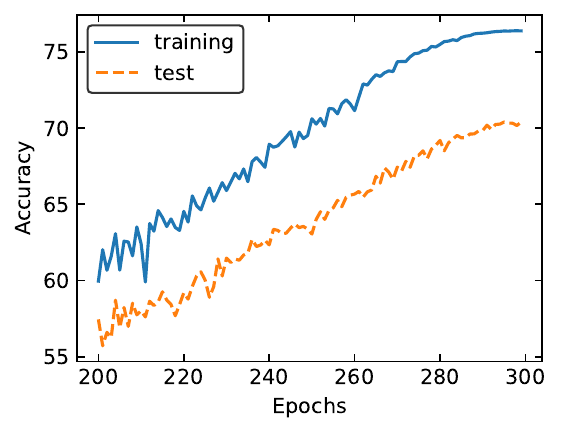}
        \caption{ResNet-18}
    \end{subfigure}
    \begin{subfigure}[h]{0.15\textwidth}
        \includegraphics[width=\linewidth, clip, trim= 0 0 0 0]{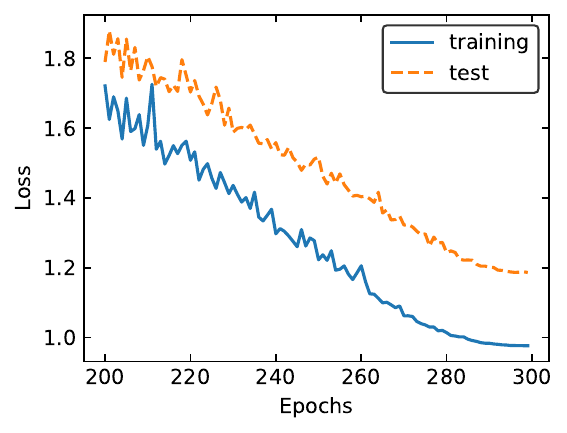}
        \caption{ResNet-18}
    \end{subfigure}
    \begin{subfigure}[h]{0.15\textwidth}
        \includegraphics[width=\linewidth, clip, trim= 0 0 0 0]{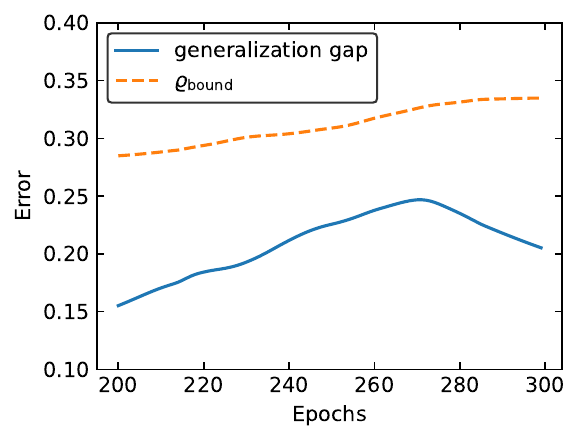}
        \caption{ResNet-18}
    \end{subfigure}\\
    \begin{subfigure}[h]{0.15\textwidth}
        \includegraphics[width=\linewidth, clip, trim= 0 0 0 0]{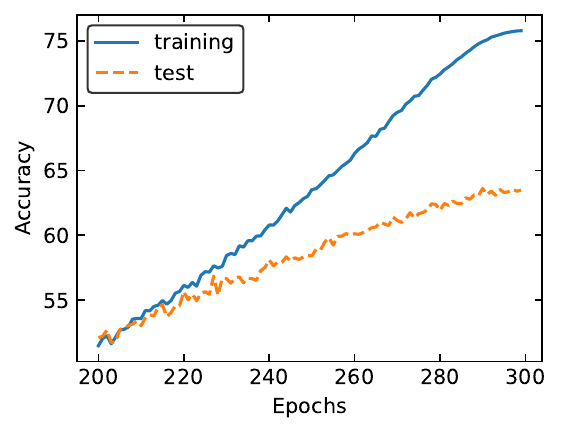}
        \caption{ViT-S-32}
    \end{subfigure}
    \begin{subfigure}[h]{0.15\textwidth}
        \includegraphics[width=\linewidth, clip, trim= 0 0 0 0]{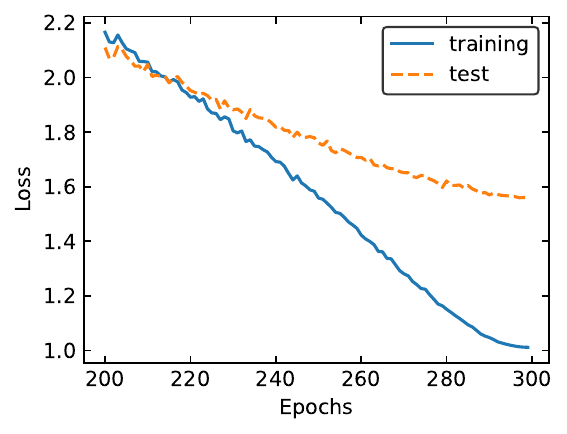}
        \caption{ViT-S-32}
    \end{subfigure}
    \begin{subfigure}[h]{0.15\textwidth}
        \includegraphics[width=\linewidth, clip, trim= 0 0 0 0]{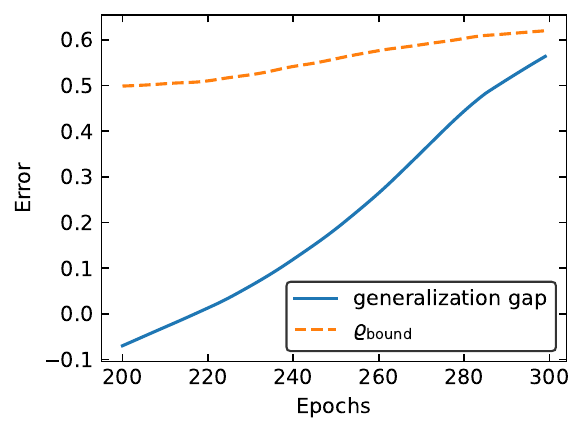}
        \caption{ViT-S-32}
    \end{subfigure}
    \caption{Evolution of different metrics for ImageNet-1K at the last 100 epochs of the training curve. Both models are trained up to 300 epochs using AdamW with the same hyperparameters.}
    \label{fig:imagenet}
\end{figure}

As shown in Fig. \ref{fig:imagenet}, we can observe that the upper bound $\varrho_{\mathrm{bound}}$ monotonically grows as a function of the training epoch, which is reasonable since the training loss keeps decreasing and as a result, the estimated diameter changes in the opposite direction.
We further note that for ResNet-18 the maximum generalization gap (\emph{i.e.}, test loss - training loss) is 0.25, and the final upper bound $\varrho_{\mathrm{bound}}$ is 0.33.
Moreover, the maximum generalization gap for ViT-S-32 is 0.5, while the final upper bound is 0.6.
It clearly shows that our estimator indeed upper-bounds the true generalization error. While these values are relatively loose at early stages of training, they become very close at late stages.
Therefore, our approach is feasible to yield plausible generalization guarantees even for such a large dataset.
\subsection{Comparison with Existing Estimators}
In this section, we quantitatively compare the Hausdorff dimension $\dim_{\text{haus}} \Gamma_{H}([0, T])$ estimated according to Eq. \eqref{eq: estimate hausdorff dimension} against other methods such as through the upper Blumenthal-Getoor index \cite{simsekli2020hausdorff} and the persistent homology dimension \cite{birdal2021intrinsic, dupuis2023generalization}.
Theoretically, these measures would be smaller if the corresponding neural network enjoys a better generalization performance.
For convenience, we still probe how they change with the number of training examples.
\par
\begin{figure}[t]
    \centering
    \begin{subfigure}[h]{0.24\textwidth}
        \includegraphics[width=\linewidth, clip, trim= 0 0 0 0]{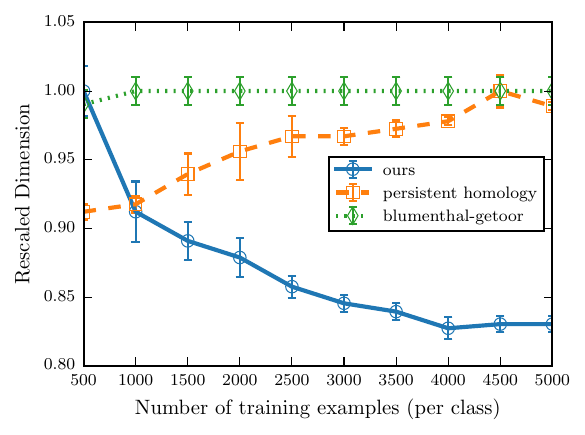}
        \caption{ResNet-56 on CIFAR-10}
    \end{subfigure}
    \hfill
    \begin{subfigure}[h]{0.24\textwidth}
        \includegraphics[width=\linewidth, clip, trim= 0 0 0 0]{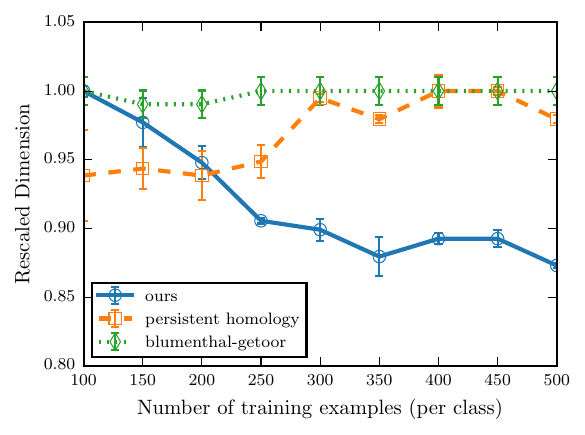}
        \caption{ResNet-56 on CIFAR-100}
    \end{subfigure}
    
    \caption{Comparison between different Hausdorff dimension estimators. For better visualization, all estimators are rescaled to the range of $[0, 1]$.}
    \label{fig:comparison of indicators}
\end{figure}

As illustrated in Fig. \ref{fig:comparison of indicators}, the persistent homology dimension increases with the training set size, which is undesirable because training with more examples generally yields better generalization.
Meanwhile, the upper Blumenthal-Getoor index stays around $1.0$ and fails to convey any information about the training set size.
By contrast, our method suggests that the Hausdorff dimension decreases with the number of training examples, which is more consistent with the true generalization error.
\section{Conclusion}
\label{sec: conclusion}
In this study, we developed a non-vacuous and tractable generalization bound for SGD from the perspective of fractal geometry, which is different from the classical generalization bounds.
Empirical results further demonstrated its efficacy by altering the training set size and the ratio of the learning rate to the mini-batch size.
Following this line, it is natural to extend our results to encompass the adaptive optimizers such as Adam and RMSprop, which we leave for future study.
\ifCLASSOPTIONcompsoc
  \section*{Acknowledgments}
\else
  \section*{Acknowledgment}
\fi
This work was supported in part by the National Natural Science Foundation of China under Grants 12501710, 12301656, 62276208, 12326607 and 12371512, in part by the Natural Science Basic Research Program of Shaanxi Province under Grant 2024JC-JCQN-02, and in part by Fundamental Research Funds for the Central Universities.
\appendix
In this section, we provide the details to prove Theorem \ref{theorem: convergence of sgd to sde} and Lemma \ref{theorem:single-trajectory}.
\begin{proof}[Proof of Theorem \ref{theorem: convergence of sgd to sde}]
According to the Hölder continuity property of FBM \cite[Theorem 1.6.1]{biagini2008stochastic}, for any $\alpha < H$, there exists a constant $C_{\omega}$ (depending on the sample path $\omega$) such that
\begin{equation*}
    |\Gamma_H(t) - \Gamma_H(s)| \leq C_{\omega} |t-s|^\alpha, \forall t, s\in[0, T].
\end{equation*}
Define the continuous-time interpolation of the discrete iterates as $\widehat{w}_t$. For $t \in [t_k, t_{k+1}]$, the discrete update can be written in integral form,
\begin{equation*}
    \widehat{w}_t = w_k - \int_{t_k}^t \mu(w_k, t_k) \mathrm{d}s + \sigma (\Gamma_H(t) - \Gamma_H(t_k)).
\end{equation*}
Subtracting this from the SDE \eqref{eq: solution to fbm sde}, the error $e_t = w_t - \widehat{w}_t$ is
\begin{equation*}
    e_t = e_{t_k} - \int_{t_k}^t (\mu(w_s, s) - \mu(w_k, t_k)) \mathrm{d}s.
\end{equation*}
Notice that the noise terms $\sigma \Gamma_H$ cancel out because the discrete scheme uses the exact increments of the noise.
Since the drift term $\mu(w_t, t)$ is Lipschitz continuous, without loss of generality, assume that there exists $M>0$ such that
\begin{equation*}
    |\mu(w_t, t) - \mu(w_s, s)| \leq M |w_t - w_s|.
\end{equation*}
Consequently, it follows that
\begin{equation*}
    |e_t| \leq |e_{t_k}| + M \int_{t_k}^t |w_s - w_k| \mathrm{d}s.
\end{equation*}
To bound $|w_s - w_k|$, look at the SDE between $t_k$ and $s$,
\begin{equation*}
    w_s - w_k = -\int_{t_k}^s \mu(w_u, u) \mathrm{d}u + \sigma (\Gamma_H(s) - \Gamma_H(t_k)).
\end{equation*}
Using the triangle inequality and the Hölder property $|\Gamma_H(s) - \Gamma_H(t_k)| \leq C_{\omega} |s-t_k|^\alpha$, it yields
\begin{equation*}
    |w_s - w_k| \leq \underbrace{\int_{t_k}^s |\mu(w_u, u)| \mathrm{d}u}_{O(\eta)} + \underbrace{\sigma C_{\omega} |s-t_k|^\alpha}_{O(\eta^\alpha)}.
\end{equation*}
Since $\alpha < H < 1$, the noise term $O(\eta^\alpha)$ dominates the drift term $O(\eta)$. Thus, there exists $C^\prime_{\omega} > 0$ such that 
\begin{equation*}
    |w_s - w_k| \leq C^\prime_{\omega} \eta^\alpha.
\end{equation*}
Now, substitute the local bound back into the error equation for $e_t$,
\begin{equation*}
    |e_t| \leq |e_{t_k}| + M \int_{t_k}^t |e_s + \underbrace{\widehat{w}_s - w_k}_{O(\eta^\alpha)}| \mathrm{d}s,
\end{equation*}
and sum these up over all steps from $0$ to $T$,
\begin{equation*}
    |e_T| \leq M \int_0^T |e_s| \mathrm{d}s + \sum_{k=0}^{K-1} M \int_{t_k}^{t_{k+1}} |\widehat{w}_s - w_k| \mathrm{d}s.
\end{equation*}
The second term is a sum of $K$ terms of order $\eta \cdot \eta^\alpha = \eta^{1+\alpha}$. Since $K = T/\eta$, the sum is $O(\eta^\alpha)$.
Applying Gronwall's Lemma, it follows that
\begin{equation*}
    |e_T| \leq (C^\prime_{\omega}MT \cdot \eta^\alpha) \exp(MT).
\end{equation*}
Taking expectations on both sides, the constant $C=\mathbb{E}[C^\prime_{\omega}]MT\exp(MT)$ is finite due to the Fernique theorem \cite[Theorem 2.7]{da2014stochastic} that an arbitrary symmetric Gaussian measure on a separable Banach space has all moments finite.
As $\eta \to 0$, the error vanishes with order $\eta^\alpha$, completing the proof.
  \end{proof}
\begin{proof}[Proof of Lemma \ref{theorem:single-trajectory}]
    Fix $r_k = \mathrm{diam}(\mathcal{H})/2^k$ and $\widehat{r}_k=\mathrm{diam}(\mathcal{H})/2^k\sqrt{m}L$.
    Then, for any $w$, $w^\prime\in \mathcal{W}$ satisfying  $\|w - w^\prime\|\leq \widehat{r}_k$, we always have for the corresponding $h_w$, $h_{w^\prime}\in \mathcal{H}$ the following inequality
    \begin{align*}
        \|h_w - h_{w^\prime}\| &= \left({\sum_{i=1}^{m} |g_w(z_i) - g_{w^\prime}(z_i)|^2}\right)^{1/2} \\
        & \leq \left({mL^2 \|w - w^\prime\|^2}\right)^{1/2} \leq r_k,
    \end{align*}
    implying that $\mathcal{N}_{r_k}(\mathcal{H}) \leq \mathcal{N}_{\widehat{r}_k}(\mathcal{W})$.
    \par
    According to Assumption \ref{assumption: regularity},  we know that $\mathcal{W}$ is regular enough so that $\dim_{\text{box}}\mathcal{W}=\dim_{\text{haus}}\mathcal{W}$.
    This means that, when $\widehat{r}_k$ approaches to zero, we have
    \begin{equation*}
        \dim_{\text{haus}}\mathcal{W} = \dim_{\text{box}}\mathcal{W} =  \lim\limits_{\widehat{r}_k\to 0} \frac{\mathcal{N}_{\widehat{r}_k}(\mathcal{W})}{-\log \widehat{r}_k}.
    \end{equation*}
    Therefore, for any $\varepsilon > 0$, there always exists an integer $k_{\varepsilon}$ such that for any $k \geq k_\varepsilon$
    \begin{equation*}
        \log \mathcal{N}_{\widehat{r}_k}(\mathcal{W}) \leq (\dim_{\text{haus}}\mathcal{W} + \varepsilon) (-\log \widehat{r}_k).
    \end{equation*}
    Choosing $c=\max(\frac{\mathcal{N}_{\widehat{r}_1}(\mathcal{W})}{(\widehat{r}_1)^{-2\varepsilon}}, \ldots, \frac{\mathcal{N}_{\widehat{r}_{k_\varepsilon}}(\mathcal{W})}{(\widehat{r}_{k_\varepsilon})^{-2\varepsilon}}, 1)$ and $\varepsilon = \dim_{\text{haus}}\mathcal{W}$, then we have for all $k\in \mathbb{N}_+$
    \begin{equation*}
        \mathcal{N}_{\widehat{r}_k}(\mathcal{W})  \leq c (\widehat{r}_k)^{-2\dim_{\text{haus}}\mathcal{W}}.
    \end{equation*}
    Substituting  $\widehat{r}_k$ in, yielding
    \begin{equation*}
        \begin{aligned}
            \log \mathcal{N}_{\widehat{r}_k}(\mathcal{W}) & \leq \log c + 2\dim_{\text{haus}}\mathcal{W} (k + \log \frac{\sqrt{m}L}{\mathrm{diam}(\mathcal{H})}) \\
            & \leq 2\dim_{\text{haus}}\mathcal{W} (k + \log \frac{\sqrt{m}Lc}{\mathrm{diam}(\mathcal{H})}) ).
        \end{aligned}
    \end{equation*}
    Write $\beta = \log \sqrt{m}L/\mathrm{diam}(\mathcal{H})$, we have
    \begin{equation*}
        \begin{aligned}
            \sqrt{\log \mathcal{N}_{r_k}(\mathcal{H})} &\leq \sqrt{\log \mathcal{N}_{\widehat{r}_k}(\mathcal{W})} \leq \sqrt{2\dim_{\text{haus}}\mathcal{W} (k + \beta + \log c)} \\
            &\leq \sqrt{2\dim_{\text{haus}} \mathcal{W}} (\frac{k}{2\sqrt{\beta + \log c}}  + \sqrt{\beta + \log c}),
        \end{aligned}
    \end{equation*}
    where the last inequality is due to the fact that $\sqrt{k+x} \leq \sqrt{x} + k/2\sqrt{x}$ for all $x>0$.
    By appealing to Dudley's lemma \cite[Lemma 27.5]{shalev2014understanding}, we complete the proof.
\end{proof}
\bibliography{main}
\bibliographystyle{IEEEtran}
\begin{IEEEbiography}[{\includegraphics[width=1.25in,height=1.25in,clip,keepaspectratio]{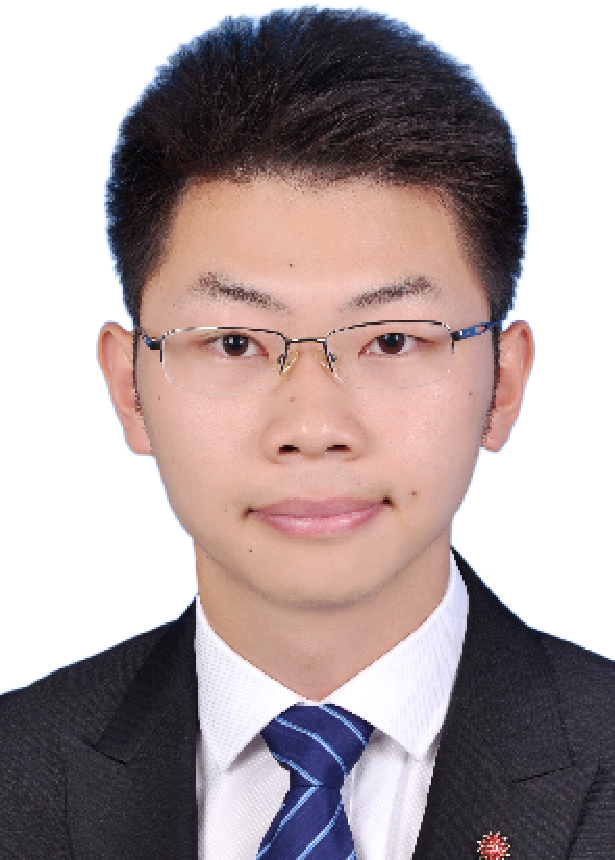}}]{Chengli Tan}
    received the M.S. and Ph.D. degrees in statistics from Xi’an Jiaotong University, Xi’an, China, in 2017 and 2024, respectively.
    Currently, he is a tenure-track associate professor at School of Mathematics and Statistics, Northwestern Polytechnical University, Xi'an, China.
    His research interests include learning theory, pattern recognition, Bayesian nonparametrics, and stochastic optimization.
\end{IEEEbiography}

\begin{IEEEbiography}[{\includegraphics[width=1.25in,height=1.5in,clip,keepaspectratio]{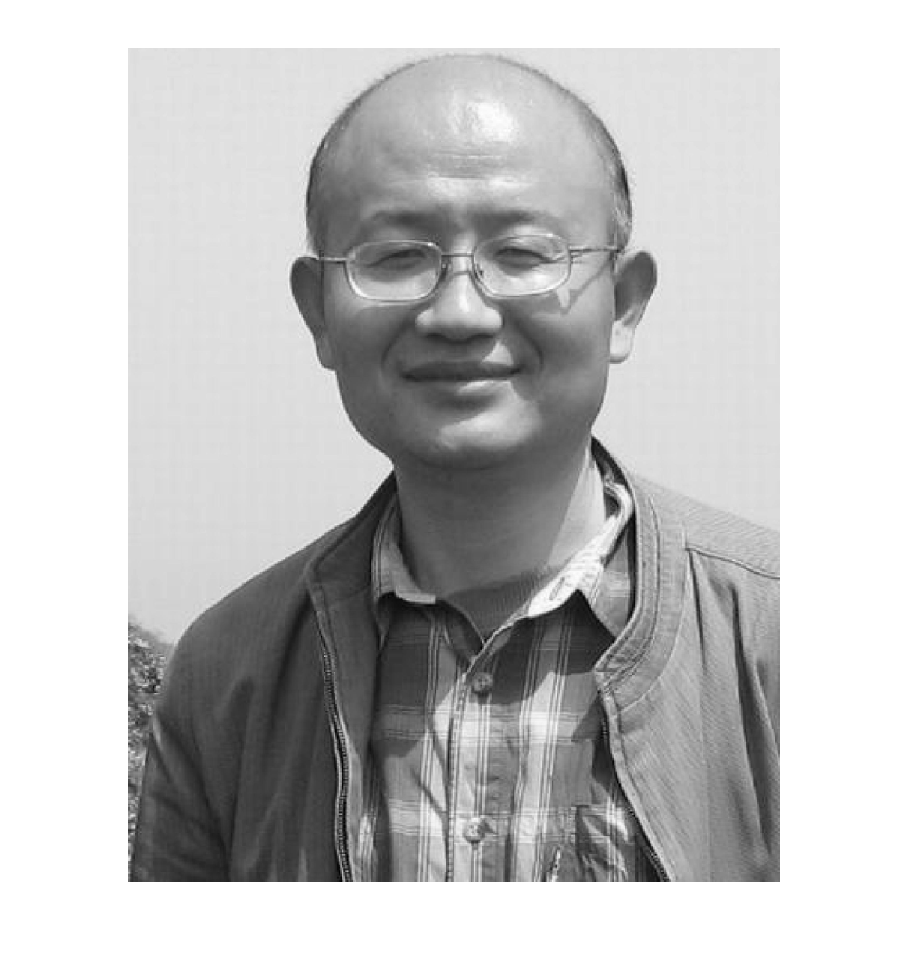}}]{Jiangshe Zhang}%
    received the B.S., M.S., and Ph.D. degrees in computational mathematics from Xi’an Jiaotong University, Xi’an, China, in 1984, 1987, and 1993, respectively. He is currently the Director of the Institute of Machine Learning and Statistical Decision Making, Xi’an Jiaotong University, where he is a Professor in the Department of Statistics. He is also the Vice-President of the Xi’an International Academy for Mathematics and Mathematical Technology, Xi’an, China. He has authored and co-authored one monograph and over 100 journal papers. His research interests include statistical computing, deep learning, cognitive representation, and statistical decision-making.
    Prof. Zhang received the National Natural Science Award of China (Third Place) and the First Prize in Natural Science from the Ministry of Education of China in 2007. He served as the President of the Shaanxi Mathematical Society and the Executive Director of the China Mathematical Society.
\end{IEEEbiography}

\begin{IEEEbiography}[{\includegraphics[width=1.25in,height=1.25in,clip,keepaspectratio]{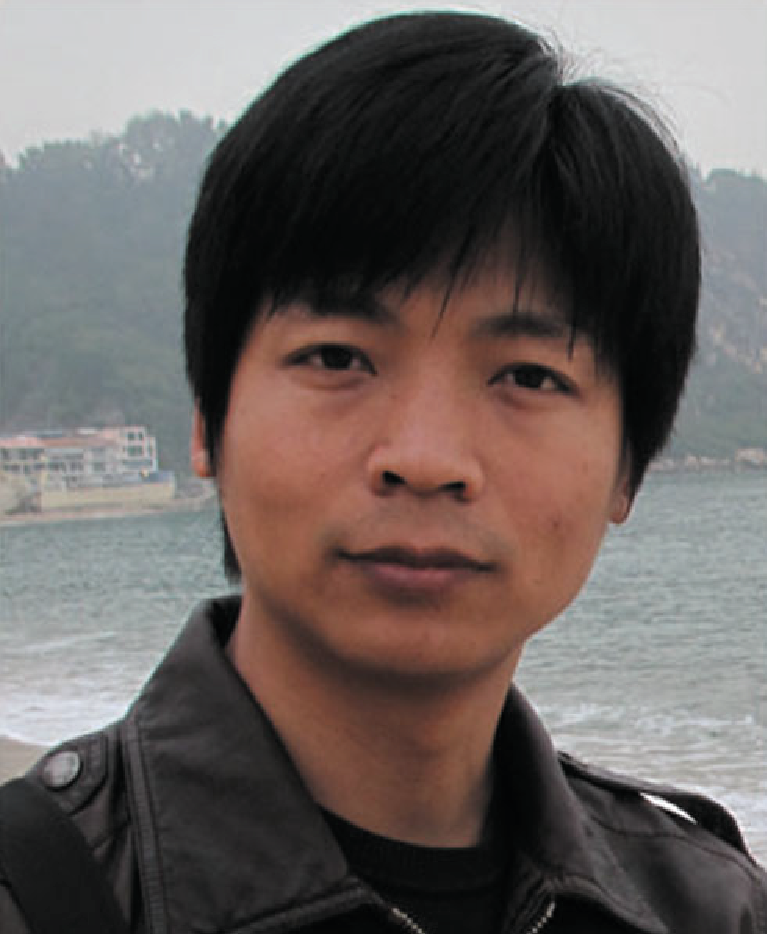}}]{Junmin Liu}%
        was born in 1982. He received his Ph.D. degree from Xi’an Jiaotong University, Xi’an, China, in 2013.
From 2011 to 2012, he served as a Research Assistant with the Department of Geography and Resource Management at the Chinese University of Hong Kong, Hong Kong, China. From 2014 to 2017, he worked as a Visiting Scholar at the University of Maryland, College Park, USA. He is currently a full Professor at the School of Mathematics and Statistics, Xi’an Jiaotong University, Xi’an, China. His research interests are mainly focused on machine learning and computer vision, typically for remotely sensed image fusion, hyperspectral unmixing, object detection, and so on. He has published over 60+ research papers in international conferences and journals.
\end{IEEEbiography}

\begin{IEEEbiography}[{\includegraphics[width=1.25in,height=1.25in,clip,keepaspectratio]{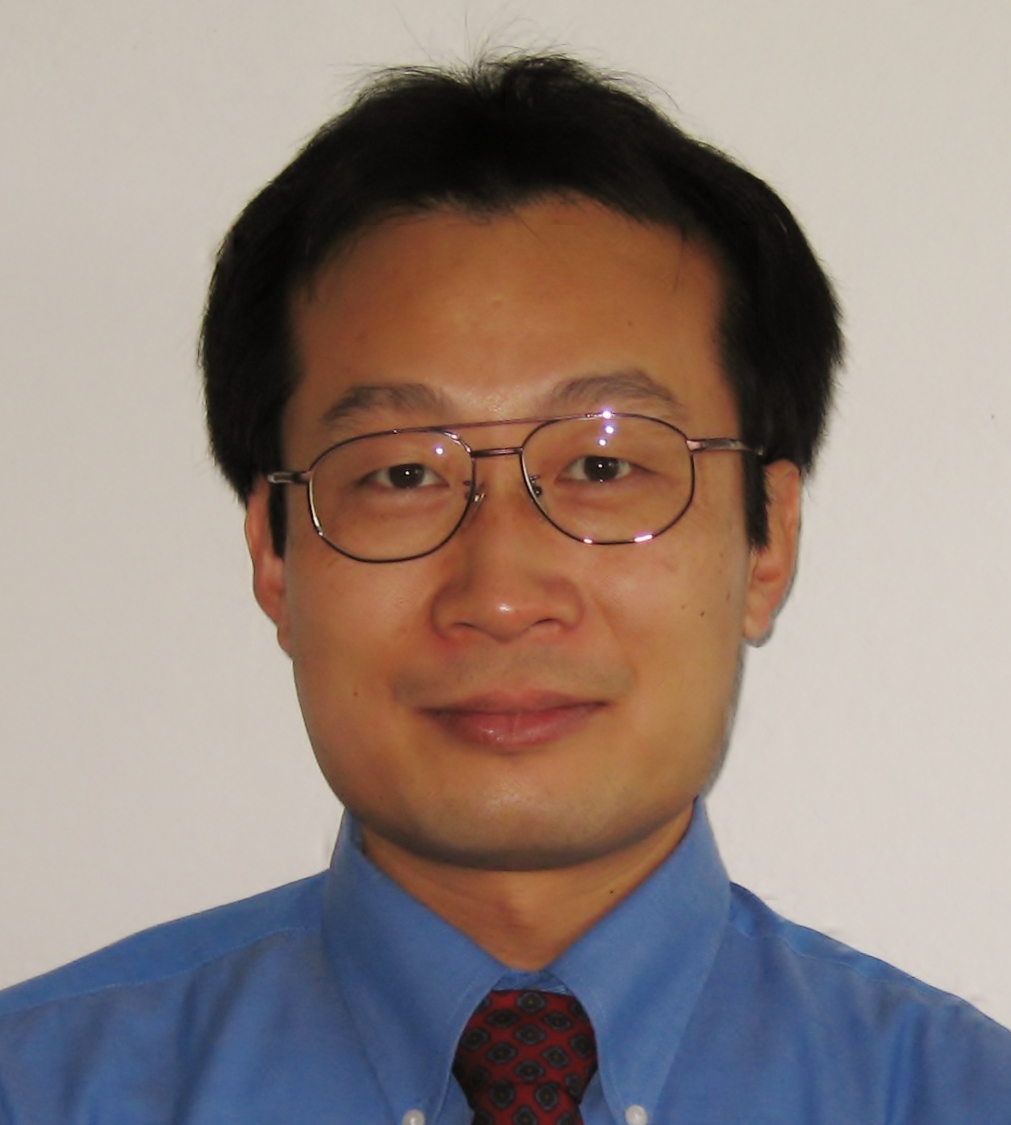}}]{Yihong Gong}%
 (Fellow, IEEE) received the B.S., M.S., and Ph.D. degrees in electrical and electronic engineering from The University of Tokyo, Tokyo, Japan, in 1987, 1989, and 1992, respectively. He was an Assistant Professor with the School of Electrical and Electronic Engineering, Nanyang Technological University, Singapore, for four years. 
 From 1996 to 1998, he was a Project Scientist with the Robotics Institute, Carnegie Mellon University, Pittsburgh, PA, USA. In 1999, he joined NEC Laboratories America, and established the Media Analytics Group for the laboratories, where he became the Site Manager to lead the entire branch with Cupertino, CA, USA. In 2012, he joined Xi’an Jiaotong University, Xi’an, China, as a Distinguished Professor. His current research interests include pattern recognition, machine learning, and multimedia content analysis.
\end{IEEEbiography}



%



\ifCLASSOPTIONcaptionsoff
  \newpage
\fi

\end{document}